\documentclass{article}
\usepackage{iclr2027_conference,times}

\setcitestyle{numbers,square,comma}

\usepackage{amsmath,amsfonts,bm}

\def\eqref#1{equation~\ref{#1}}
\def\1{\bm{1}}

\DeclareMathAlphabet{\mathsfit}{\encodingdefault}{\sfdefault}{m}{sl}
\SetMathAlphabet{\mathsfit}{bold}{\encodingdefault}{\sfdefault}{bx}{n}

\usepackage{hyperref}
\usepackage{url}
\usepackage{graphicx}
\usepackage[export]{adjustbox} % 用于图片垂直居中对齐 (valign=c)
\usepackage{subcaption}        % 用于美化多子图排版
\usepackage{amsmath,amssymb}
\usepackage{pifont}            % 用于标准勾叉符号
\usepackage{amsthm}
\usepackage{placeins} 

\newcommand{\cmark}{\textcolor{green!60!black}{\ding{51}}}
\newcommand{\xmark}{\textcolor{red!75!black}{\ding{55}}}

\usepackage[table]{xcolor}   
\usepackage{colortbl}
\usepackage{booktabs}
\usepackage{multirow}

\definecolor{headblue}{RGB}{219, 234, 245}  % 表头淡蓝灰底色
\definecolor{rowours}{RGB}{228, 236, 243}   % 本文方法整行的淡冷蓝灰底色
\definecolor{cellorange}{RGB}{254, 235, 226}% 分组1末列/指标高亮淡橙色
\definecolor{cellyellow}{RGB}{254, 249, 219}% 分组2末列/指标高亮淡黄色
\definecolor{cellgreen}{RGB}{186, 247, 187} % 本文方法最强指标强调浅绿色

\newcommand{\cHB}{\cellcolor{headblue}}
\newcommand{\cRO}{\rowcolor{rowours}}
\newcommand{\cCO}{\cellcolor{cellorange}}
\newcommand{\cCY}{\cellcolor{cellyellow}}
\newcommand{\cCG}{\cellcolor{cellgreen}}

\newcommand{\nobooktabssep}{%
  \setlength{\aboverulesep}{0pt}%
  \setlength{\belowrulesep}{0pt}%
}

\title{Representation--Behavior Alignment for Explainable Weakly-Supervised Video Anomaly Detection}

\author{%
  \textbf{Chao Huang}\textsuperscript{1} \quad
  \textbf{Pengfei Wei}\textsuperscript{1} \quad
  \textbf{Kaige Li}\textsuperscript{1} \quad
  \textbf{Chengliang Liu}\textsuperscript{2} \\[5pt]
  \textbf{Wei Wang}\textsuperscript{1} \quad
  \textbf{Wenqi Ren}\textsuperscript{1} \quad
  \textbf{Xiaochun Cao}\textsuperscript{1} \\[8pt]
  \textsuperscript{1}Sun Yat-sen University \qquad
  \textsuperscript{2}University of Macau
}

\iclrfinalcopy

\begin{document}

\maketitle
\thispagestyle{fancy}
\lhead{Preprint. Under review.}

\begin{abstract}
Multimodal Large Language Models (MLLMs) provide a natural way to make video anomaly detection more explainable. However, their final decisions do not always fully use the discriminative information contained in their hidden states, an issue we refer to as representation--behavior misalignment. We decompose this gap into a capacity component that measures discriminative information never aggregated into the readout position, and a directional component that measures the angular mismatch between the optimal and the native normal--abnormal axis at that position. Across multiple video anomaly detection benchmarks and MLLM backbones the directional component dominates, and residual-stream tracing shows that native-axis separability rises sharply in several mid-to-late attention layers. Because both components are governed by attention rather than MLP updates, we propose Representation--Behavior Alignment (RBA), a parameter-efficient method that adapts those layers using video-level labels alone while updating about 0.012\% of the backbone parameters. Experiments on three benchmarks show that RBA improves native-readout performance and better aligns the model's decision direction with discriminative representations, and it produces anomaly decisions and explanations through a single generative process.
\end{abstract}

% -------------------------------------------------------------
% 1. Introduction
% -------------------------------------------------------------
\section{Introduction}
\label{sec:intro}

Multimodal Large Language Models (MLLMs) unify visual understanding, decision making, and natural language explanation in one interface. Yet the decisions they generate do not necessarily reflect the task-relevant information their internal representations already hold. Work on linear probing and latent prediction has shown that frozen hidden states often encode task signals the output interface cannot reliably express~\citep{alain2016probes,belinkov2022probing,burns2022latent,belrose2023tunedlens}. We examine this representation--behavior gap in weakly-supervised Video Anomaly Detection (VAD), asking how much of the performance loss is tied to the native readout direction, where native-readout alignment emerges across depth, and whether one targeted parameter-efficient intervention can mitigate it.

\paragraph{Why Video Anomaly Detection.}
VAD suits this question well. Judging a clip as anomalous depends on how a short span of frames evolves and on the surrounding scene, so no single frame settles it. Training data typically carry one label per video, yet a useful system must localize anomalies and explain its judgment in language. Existing methods cover only part of this need. Weakly-supervised detectors~\citep{tian2021weakly,wu2024vadclip} are accurate but sit outside the language-generation pathway, training-free MLLMs explain their outputs yet lack precision, and instruction-tuned MLLMs~\citep{zhang2025holmes} lean on costly frame annotations or reasoning traces. We fix a four-frame clip window throughout. That caps the absolute accuracy any readout can reach, but it applies identically to every readout we compare, so the gaps we report are internal to the model rather than a property of the input window.

\begin{figure}[!t]
\centering
\begin{minipage}[c]{0.58\textwidth}
    \centering
    \includegraphics[width=\linewidth]{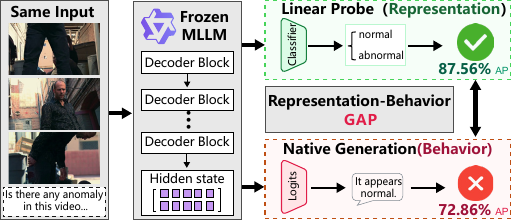}
\end{minipage}%
\hfill
\begin{minipage}[c]{0.38\textwidth}
    \centering
    \includegraphics[width=\linewidth]{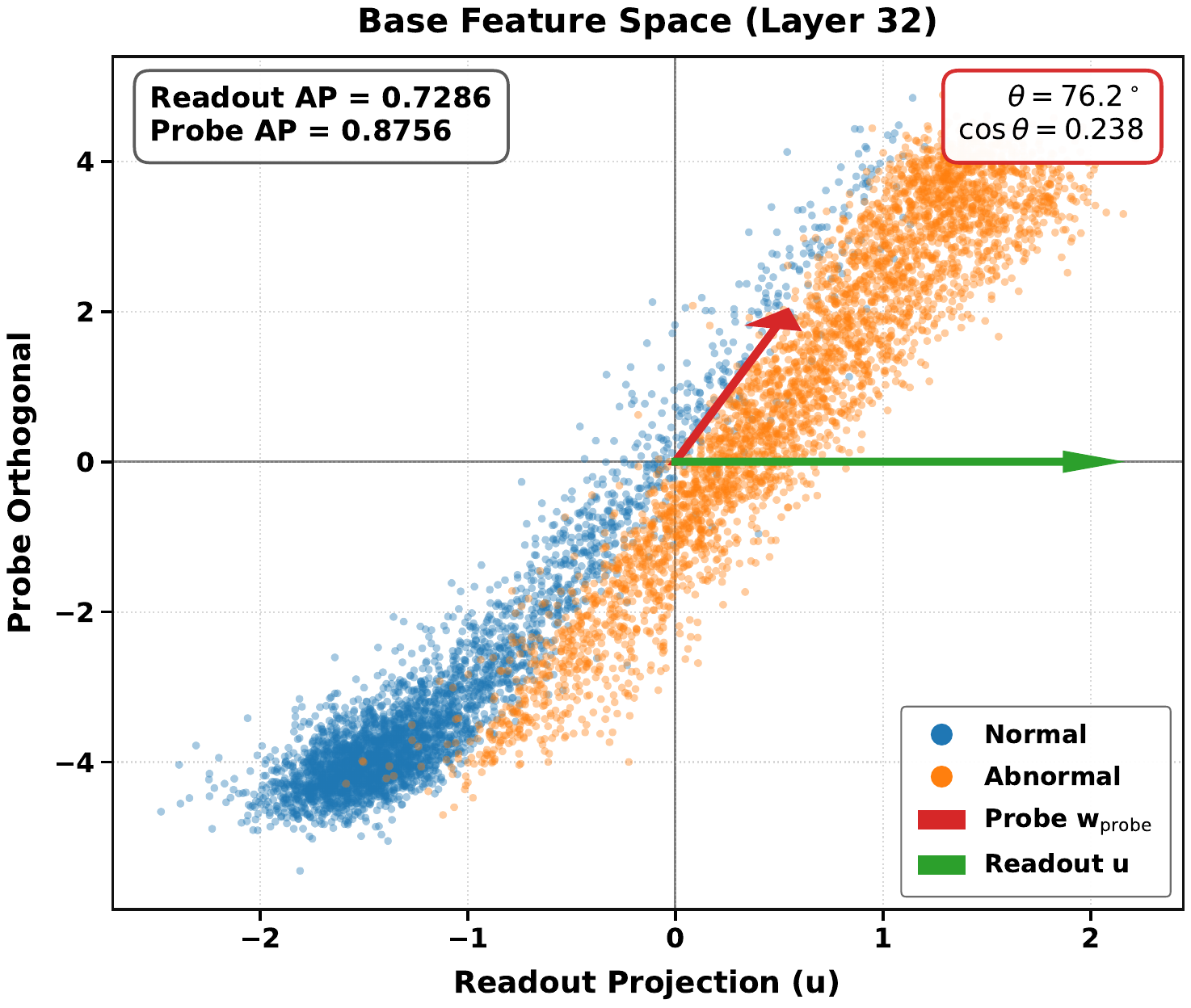}
\end{minipage}

\vspace{3pt}

\begin{minipage}[t]{0.58\textwidth}
    \caption*{(a) Representation--Behavior Gap}
    \label{fig:gap}
\end{minipage}%
\hfill
\begin{minipage}[t]{0.38\textwidth}
    \caption*{(b) Directional Misalignment}
    \label{fig:misalignment}
\end{minipage}

\vspace{-4pt}

\caption{Diagnosis of representation--behavior misalignment on XD-Violence. (a) Native readout versus one-dimensional readouts on frozen hidden states. (b) 2D projection showing directional misalignment between the Fisher-optimal axis $\mathbf{w}_{\mathrm{LDA}}$ and the native readout axis $\mathbf{u}$.}
\label{fig:overview}
\end{figure}

\paragraph{Core Diagnosis.}
Evaluating the frozen final-layer features under identical inputs reveals a clear divergence between decision pathways. The native readout reaches only 0.7286 Frame-AP on XD-Violence, whereas the best one-dimensional readout on the same state attains 0.9588 and a weakly supervised linear probe already recovers 0.8756, as depicted in Figure~\ref{fig:overview}(a). Chaining two controlled comparisons attributes 97.5\% of that gap to the readout direction rather than to information missing from the readout position. Null-space projections agree, since stripping out the frame-supervised direction $\mathbf{w}_{\mathrm{LDA}}$ degrades accuracy substantially. Figure~\ref{fig:overview}(b) shows this optimal direction sitting $76.2^\circ$ away from the native readout axis $\mathbf{u}$, a cosine similarity of only 0.238.

Tracking how well the residual stream separates normal from abnormal clips along the native direction locates where that alignment appears. It emerges abruptly at an attention sublayer in the middle-to-late network. Separability more than triples at Block~20 and then levels off, with attention updates improving the alignment and MLP updates weakening it. The probe, meanwhile, is already near its final performance before the transition. Useful anomaly features thus form early, but the model connects them to its language readout late and only partially.

\paragraph{Method and Contributions.}
We propose Representation--Behavior Alignment (RBA) in response. RBA adds only about 0.012\% trainable parameters to the attention sublayers pinpointed by the analysis, and uses pseudo-labels from the probe to push representations toward the native readout direction. A single forward pass at test time then yields anomaly predictions together with explanations that match those decisions. Our contributions can be summarized as follows.

\begin{itemize}
    \item \textbf{Geometric diagnosis for VAD.} Comparing readouts that differ in one constraint at a time shows that projection direction, not the information available at the readout position, accounts for nearly all of the observed performance gap. Separating the actionable directional term from the capacity term reveals that the missing performance reflects a misoriented native axis rather than uninformative features.
    
    \item \textbf{Layer-wise attribution.} We locate the sharp change in native-axis separability at specific attention sublayers and distinguish the roles of the attention and MLP modules. This attribution turns a diffuse ``the model underperforms'' observation into a data-driven criterion for which layers to adapt.
    
    \item \textbf{A diagnosis-guided, lightweight intervention.} With video-level labels alone and 0.012\% trainable parameters, RBA rotates the discriminative direction toward the native readout axis at the identified layers. It delivers consistent gains across three VAD benchmarks and several MLLM backbones while preserving native decision explainability.
\end{itemize}

% -------------------------------------------------------------
% 2. Related Work
% -------------------------------------------------------------
\section{Related Work}
\label{sec:related_work}

\paragraph{Reading Internal Representations.}
Linear probing~\citep{alain2016probes,belinkov2022probing}, together with unembedding-projection tools such as Logit Lens~\citep{belrose2023tunedlens,nostalgebraist2020logitlens} and Representation Engineering~\citep{zou2023representation}, has shown that the internal representations of language models often encode more task-relevant information than their explicit textual outputs reveal. Most of this work, however, remains at the level of passive diagnostic observation. While these studies confirm that such a performance discrepancy exists, they pay less attention to geometrically decomposing the gap into direction and capacity components, to localizing where it emerges across network depth, or to deriving targeted parameter-efficient interventions from those findings. In this work, we extend passive inspection into an actionable pipeline that covers diagnosis, layer-wise attribution, and a minimal intervention for MLLM-based decision tasks.

\paragraph{Video Anomaly Detection.}
Existing Video Anomaly Detection (VAD) paradigms fall broadly into two categories. Weakly-supervised specialized detectors~\citep{tian2021weakly,wu2024vadclip}, typically built upon Multiple Instance Learning (MIL), achieve strong detection accuracy on standard benchmarks. Their scalar predictions, however, bypass the generative language pathway and decouple textual explanations from the detection mechanism. In contrast, MLLM-based VAD frameworks comprise training-free inference~\citep{zanella2024harnessing} and instruction tuning~\citep{zhang2025holmes}. Training-free approaches unify judgment and explanation but suffer from lower precision due to readout misalignment. Instruction tuning improves accuracy yet demands dense frame annotations or reasoning chains that conflict with realistic weakly-supervised settings. Our proposed RBA reformulates this challenge as a directional alignment problem along the native readout axis. Supervised solely by video-level labels, it performs minimal low-rank adaptation to deliver accurate detection and consistent explanations with no external inference classifier.
% -------------------------------------------------------------
% 3. Diagnosis and Attribution
% -------------------------------------------------------------
\section{Representation--Behavior Misalignment: Diagnosis and Attribution}
\label{sec:diagnosis}

Hidden states in MLLM-based VAD can carry discriminative information that the generated answer never uses, a shortfall we call \emph{representation--behavior misalignment}. To measure it, we compare readouts that differ in exactly one constraint at a time, namely how much representational capacity a readout may use and which direction it must read along. This yields a two-term accounting of the gap, and residual-stream tracing then locates the depth at which each term is resolved.

\subsection{Formalization and Two-Term Gap Decomposition}
\label{sec:formulation_and_decomposition}

\paragraph{Task Setup and Readout Position.}
For a video $\mathcal{V}$, each clip $i$ is placed in a fixed prompt template, so every clip yields a token sequence of the same length $N$ built from prefix tokens, the visual tokens of four frames, and the question tokens. Let $\tilde{\mathbf{h}}^{(p)}_{L,i} \in \mathbb{R}^{d}$ denote the RMSNorm-normalized final-layer residual state at position $p$~\citep{zhang2019rmsnorm}, the state produced after the last MLP update and read by the output head. We read at the final position $p = N$, a single prompt token, and abbreviate $\tilde{\mathbf{h}}_{L,i} := \tilde{\mathbf{h}}^{(N)}_{L,i}$. We write $\mathbf{a}_{l,i}$ for the attention-sublayer output of block $l$ at that position, the only update in the block able to aggregate information from other token positions into it. The model's \emph{native readout axis} $\mathbf{u}$ is the unit direction in the pre-trained output embedding matrix pointing from the ``Normal'' token to the ``Abnormal'' token,
\begin{equation}
\mathbf{u} = \frac{\mathbf{W}_U[\text{Abnormal}] - \mathbf{W}_U[\text{Normal}]}{\|\mathbf{W}_U[\text{Abnormal}] - \mathbf{W}_U[\text{Normal}]\|_2},
\end{equation}
and the projection $d_i = \mathbf{u}^\top \tilde{\mathbf{h}}_{L,i}$ is the native abnormal-versus-normal score of clip $i$. Every readout below shares the same prompt, clips, visual input, and network weights, so only the available positions and the reading direction vary. Properties of the input itself, including the temporal window length, are held fixed and cannot account for differences among them.

\paragraph{Three Readout Tiers.}
The \emph{sequence-level reference} $P_{\mathrm{full}}$ fits linear discriminant analysis on the concatenation of final-layer states at \emph{every} position,
\begin{equation}
\mathbf{g}_{L,i} = \big[\tilde{\mathbf{h}}^{(1)}_{L,i};\, \tilde{\mathbf{h}}^{(2)}_{L,i};\, \dots;\, \tilde{\mathbf{h}}^{(N)}_{L,i}\big] \in \mathbb{R}^{Nd},
\end{equation}
exposing the visual and prompt tokens alongside the readout token. It bounds the linearly extractable signal in the whole sequence whether or not the model has routed that signal into the readout token, though not the nonlinear information the sequence may also hold. The \emph{position-constrained reference} $P_{\mathrm{last}}$ fits Fisher LDA on the single last-position state with closed-form optimum $\mathbf{w}_{\mathrm{LDA}} = \boldsymbol{\Sigma}_w^{-1}(\boldsymbol{\mu}_1 - \boldsymbol{\mu}_0)$, the best one-dimensional readout once the signal has been compressed into the one token the output head consumes. The \emph{native generation readout} $P_{\mathrm{read}}$ projects that same state onto the fixed direction $\mathbf{u}$. Both references use held-out frame annotations retrospectively, so they are diagnostic quantities rather than attainable operating points.

\paragraph{Decomposition.}
Comparing $P_{\mathrm{full}}$ with $P_{\mathrm{last}}$ varies the available positions while both sides use the Fisher-optimal direction, and comparing $P_{\mathrm{last}}$ with $P_{\mathrm{read}}$ fixes the position and varies only the direction. Chaining the two gives
\begin{equation}
\label{eq:decomposition}
\Delta_{\mathrm{total}} = P_{\mathrm{full}} - P_{\mathrm{read}}
= \underbrace{\big(P_{\mathrm{full}} - P_{\mathrm{last}}\big)}_{\Delta_{\mathrm{cap}}\ \text{(Capacity / Aggregation Term)}}
+ \underbrace{\big(P_{\mathrm{last}} - P_{\mathrm{read}}\big)}_{\Delta_{\mathrm{dir}}\ \text{(Directional Term)}},
\end{equation}
where $\Delta_{\mathrm{cap}}$ measures discriminative information present somewhere in the sequence, notably in the visual tokens, yet unrecoverable from the readout position alone, and $\Delta_{\mathrm{dir}}$ measures the cost of reading that position along $\mathbf{u}$ rather than $\mathbf{w}_{\mathrm{LDA}}$. Equation~\ref{eq:decomposition} is exact by construction and claims no statistical independence between the terms. Its value lies in the controls it enforces, since each term isolates one varied constraint while the shared intermediate $P_{\mathrm{last}}$ holds the other fixed.

\paragraph{The Probe as an Estimator.}
Frame labels are unavailable at training time, so $\mathbf{w}_{\mathrm{LDA}}$ cannot be computed in the deployed setting. We therefore fit a Top-$K$ Multiple Instance Learning (MIL) linear probe at the readout position as its weakly supervised estimator using only the video-level label $Y_j$,
\begin{equation}
\min_{\mathbf{w}, b} \sum_j \mathcal{L}_{\mathrm{BCE}}\Big(Y_j, \operatorname*{pool}_{i \in \text{Top-}K} \sigma(\mathbf{w}^\top \tilde{\mathbf{h}}_{L,i} + b)\Big).
\end{equation}
The probe is not a tier in Equation~\ref{eq:decomposition}. We judge it by its fidelity to the Fisher optimum, reported as $\cos(\mathbf{w}_{\mathrm{probe}}, \mathbf{w}_{\mathrm{LDA}})$ and $\varepsilon_{\mathrm{probe}} = P_{\mathrm{last}} - P_{\mathrm{probe}}$, and it only produces offline pseudo-labels rather than being retained at inference. Because $\mathbf{g}_L$ has $N$ times the dimension of $\tilde{\mathbf{h}}_L$, we reduce it by PCA fitted on the training split, estimate $\boldsymbol{\Sigma}_w$ with Ledoit--Wolf shrinkage, and fit every discriminant on the training split alone.

\subsection{Geometric Attribution and Residual-Stream Readout Transition}
\label{sec:attribution_and_dynamics}
\label{sec:misalignment_diagnosis}

\paragraph{The Directional Term Dominates.}
On XD-Violence with Qwen3.5-9B, the sequence-level reference reaches 0.9646 Frame-AP and the position-constrained reference 0.9588, whereas the native readout obtains only 0.7286. Within the total gap of 0.2360, the directional term accounts for 0.2302, or 97.5\%, leaving 0.0058 to the capacity term. Projecting out the normalized LDA direction via $\mathbf{h}^\perp = (\mathbf{I} - \hat{\mathbf{w}}_{\mathrm{LDA}}\hat{\mathbf{w}}_{\mathrm{LDA}}^\top)\tilde{\mathbf{h}}_L$ causes a sharp drop, confirming that the linearly accessible signal concentrates along this single vector. The Fisher-optimal axis deviates from the native axis by $76.2^\circ$, a cosine similarity of 0.238, while the weakly supervised estimator tracks that optimum closely at $\cos(\mathbf{w}_{\mathrm{probe}}, \mathbf{w}_{\mathrm{LDA}}) = 0.907$. The readout position is thus not where the signal is lost. Recoverable performance is forfeited mainly to a misoriented axis.

\begin{figure}[!t]
\centering
\includegraphics[width=0.98\linewidth]{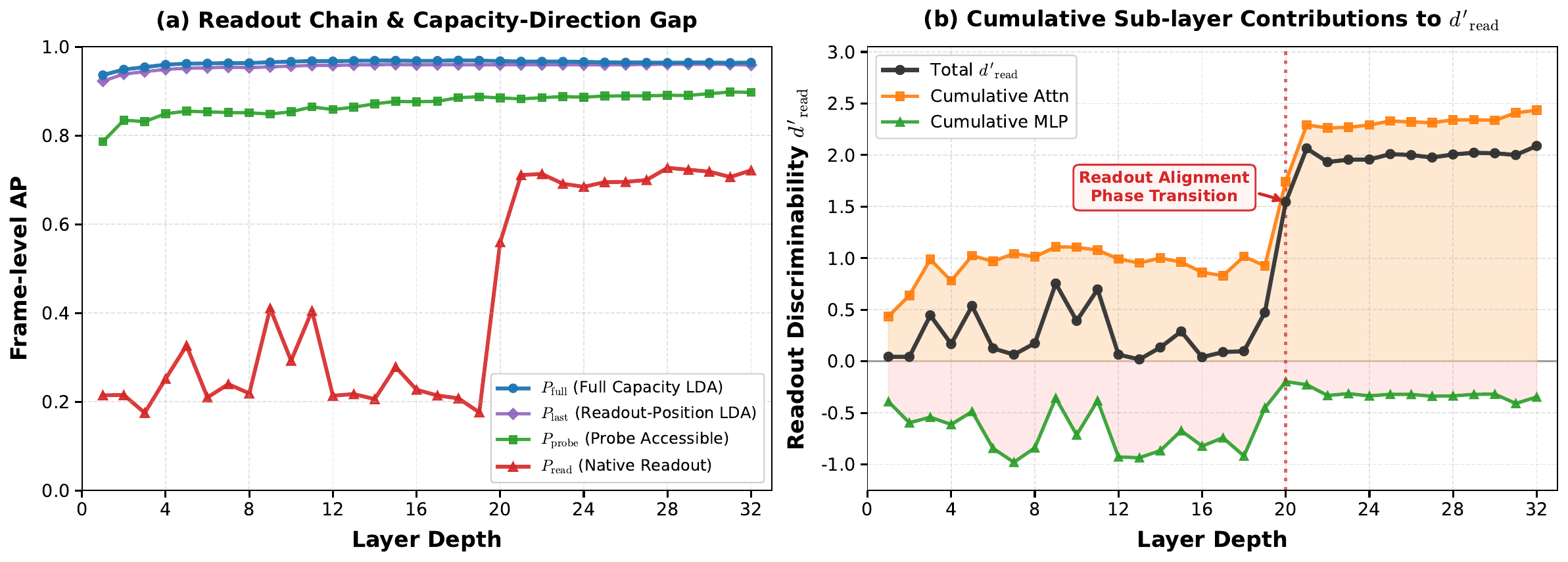}
\caption{Layer-wise residual-stream diagnostic. The two Fisher references nearly coincide at every depth and saturate early, whereas native separability increases sharply at the Block~20 attention sublayer. Additive decomposition isolates positive gains from attention and negative interference from MLP sublayers.}
\label{fig:phase_transition}
\end{figure}

\paragraph{A Localized Readout Transition and the Opposition of Sublayers.}
Sweeping Equation~\ref{eq:decomposition} over all 32 blocks shows $\Delta_{\mathrm{cap}}$ staying uniformly small and shrinking with depth, from 1.35 AP points at Block~1 to 0.35 at Block~28, so the two references in Figure~\ref{fig:phase_transition} are nearly indistinguishable and saturate within the first few layers. Separability along $\mathbf{u}$, measured at 64 sampling points in the residual stream, follows a very different trajectory. The statistic $d'_{\mathrm{read}}(l)$ jumps from 0.474 to 1.545 at the attention sublayer of Block~20, a rise of 1.07 with no gradual buildup beforehand. The representation supports anomaly discrimination long before the model can express that distinction through its own output tokens, which rules out late feature formation as an explanation.

The right panel of Figure~\ref{fig:phase_transition} accumulates each attention or MLP residual update's contribution to native-axis separability. The attention curve takes a large positive step at the transition while the MLP curve stays negative. This asymmetry matches what the two operators can do, since attention alone moves information across token positions and can therefore reduce $\Delta_{\mathrm{cap}}$ or write context-dependent content onto the readout axis, whereas MLP updates act per position and only rotate what is already there. We read the transition as a routing event, where discriminative content present elsewhere in the sequence since the early layers is written into the readout position and partially projected onto $\mathbf{u}$. This account describes the configurations we measure rather than a verified causal mechanism, and Appendix~\ref{app:mechanism} adds further depth-wise analysis.

\subsection{Diagnostically Derived Design Principles}
\label{sec:design_constraints}

Three design choices follow. First, attention is the operator implicated in both terms, since only attention moves information into the readout position and the same residual updates write the direction there, which makes attention projections at the transition our intervention point and motivates freezing the MLPs rather than leaving that choice arbitrary. Second, the signal that must be redirected occupies the small subspace spanned by $\mathbf{w}_{\mathrm{LDA}}$ and $\mathbf{u}$, so a low-rank update at the identified blocks suffices without reshaping general representations. Third, $\Delta_{\mathrm{dir}}$ is an angular quantity and anomaly detection is a ranking problem, so we optimize the margin along the fixed native direction instead of matching the probe's probability distribution.
% -------------------------------------------------------------
% 4. Methodology
% -------------------------------------------------------------
\section{Methodology: Representation--Behavior Alignment}
\label{sec:method}

\begin{figure}[!t] 
\centering
\includegraphics[width=0.96\linewidth]{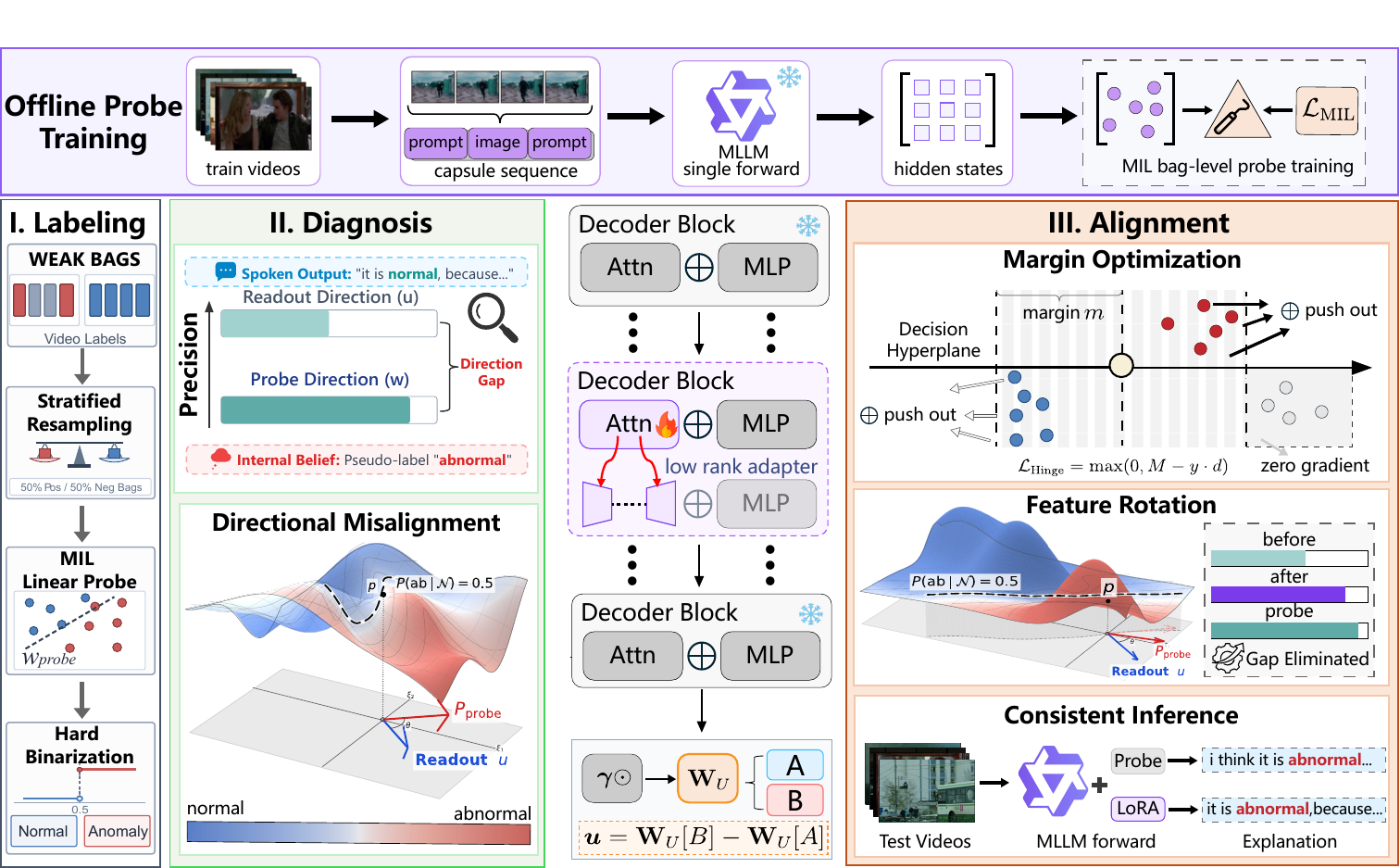}
\caption{Overview of the RBA framework. Probe-derived pseudo-labels supervise low-rank adapters injected into the transition attention layers, optimized via a hinge-margin loss along the fixed native readout axis $\mathbf{u}$.}
\label{fig:framework}
\end{figure}

The goal of RBA is to align discriminative features with the model's native generation direction under weak supervision. Figure~\ref{fig:framework} summarizes the whole procedure: we first train a probe and generate pseudo-labels, then add low-rank adapters at the transition layers, and finally optimize the native decision margin along the fixed readout axis. The diagnosis in Section~\ref{sec:attribution_and_dynamics} is run before adapter training and determines \emph{where} RBA acts, so it adds no inference-time component. RBA needs neither an external teacher nor frame annotations, and no separate classifier at inference.

\subsection{Overall Pipeline}
\label{sec:overall_pipeline}

We split each video into temporal clips and place them in a fixed prompt made up of a prefix, visual tokens, and a question. In every stage, we extract a segment representation at the prediction-token position, which keeps the representation space consistent between probe training and alignment. Guided by the residual-stream diagnosis in Section~\ref{sec:attribution_and_dynamics}, the adapters are inserted into the identified transition layer and the block that immediately follows it. In this way, the probe supplies the training signal while the residual-stream diagnosis supplies the adapter location.

\subsection{Probe-Derived Internal Supervision}
\label{sec:probe_supervision}

Let video $j$ denote an instance bag with video-level label $Y_j \in \{0, 1\}$. We train a linear probe on the normalized final hidden states using Top-$K$ MIL pooling:
\begin{equation}
\min_{\mathbf{w}, b} \sum_j \mathcal{L}_{\mathrm{BCE}}\Big(Y_j, \operatorname*{pool}_{i \in \text{Top-}K} \sigma(\mathbf{w}^\top \tilde{\mathbf{h}}_{L,i} + b)\Big).
\end{equation}
To mitigate class imbalance, we perform stratified class-balanced sampling strictly at the video level, preventing leakage between temporal segments from the same video.

After training, the probe produces a posterior score $s_i$ for each segment. We threshold this score at 0.5 to obtain the pseudo-label $\hat{y}_i$, where $-1$ denotes normal and $+1$ denotes abnormal. These labels remain fixed while training RBA. We use only their direction, rather than the probe's potentially uncalibrated probability values. Notably, the probe is not retained as an inference-time head; it provides offline supervision only.

\subsection{Targeted Low-Rank Adaptation}
\label{sec:lora_adaptation}

We add LoRA parameters~\citep{hu2021lora} only to the attention projections in the transition layers. The following equation defines the adapted weight $\mathbf{W}'$ as the frozen base weight plus a low-rank update:
\begin{equation}
\mathbf{W}' = \mathbf{W}_{\mathrm{base}} + \frac{\alpha}{r}\mathbf{B}\mathbf{A}, \qquad \mathbf{A} \in \mathbb{R}^{r \times d_{\mathrm{in}}}, \quad \mathbf{B} \in \mathbb{R}^{d_{\mathrm{out}} \times r}.
\end{equation}
We use rank $r = 16$, scaling factor $\alpha = 32$, and apply LoRA to the attention projections in zero-indexed Blocks 20--21. The MLP sublayers and all remaining backbone parameters are frozen, restricting trainable parameters to roughly 0.012\% of the backbone. The adapters can be merged into the base weights at inference without introducing any additional computational latency.

\paragraph{Low-Rank Geometric Motivation.}
Because our empirical diagnosis reveals that the critical discriminative signal occupies the low-dimensional subspace spanned by $\mathbf{w}_{\mathrm{probe}}$ and $\mathbf{u}$, updating the entire backbone or reshaping general representations is unnecessary. Instead, lightweight low-rank modifications suffice to rotate internal features toward the native axis $\mathbf{u}$. We set the LoRA rank to $r = 16$ to accommodate cross-layer dynamics alongside inter-video variance. As reported in Table~\ref{tab:gap_decomposition}, this targeted intervention substantially increases alignment with the native readout direction.

\subsection{Decision-Axis Margin Alignment Objective}
\label{sec:margin_loss}

During fine-tuning, the native decision score $d_i$ is the abnormal-minus-normal token logit difference at the first answer position, equivalent to a projection onto the fixed native readout direction up to a positive constant scale. The unembedding matrix and the final RMSNorm remain frozen, ensuring that optimization changes the hidden representations rather than the classification axis itself.

We optimize the unweighted hinge-margin loss:
\begin{equation}
\label{eq:hinge}
\mathcal{L}_{\mathrm{align}} = \frac{1}{B}\sum_{i=1}^{B} \max\big(0, M - \hat{y}_i \cdot d_i\big),
\end{equation}
where $M = 2.0$. Once a representation satisfies $\hat{y}_i d_i \geq M$, its gradient vanishes, concentrating the limited adapter capacity on samples that are incorrectly oriented or insufficiently separated.

We choose the hinge loss over cross-entropy or KL divergence because alignment calls for correct orientation and separation along the native axis, not replication of the probe's probability distribution. Compared with distribution-matching objectives, the hinge objective directly improves native-axis separation without assuming that the probe posterior is calibrated.

\subsection{Capacity Isolation Principle}
\label{sec:capacity_isolation}

If we add an explanation loss $\mathcal{L}_{\mathrm{explain}}$ to the same adapters, detection AP on XD-Violence drops from 0.8302 to 0.7279, completely erasing the alignment gain and falling slightly below the zero-shot baseline of 0.7286. This happens because the decision loss acts on a single direction, whereas the explanation loss affects many token positions and directions. When the adapter rank is only 16, the two objectives compete for capacity.

We call this the \emph{capacity isolation principle}. During fine-tuning, $\mathcal{L}_{\mathrm{align}}$ is evaluated only at the first answer position and updates only the selected attention adapters, and we do not add an explanation-generation loss. Explanations draw directly on the frozen generative ability of the pre-trained model. At inference, scores and explanations then come from the same aligned pathway.

% -------------------------------------------------------------
% 5. Experiments
% -------------------------------------------------------------
\section{Experiments}
\label{sec:experiments}

Three questions organize our study. Does the representation--behavior gap appear consistently across datasets and MLLM backbones? Can a diagnosis-guided low-rank update improve native anomaly scores from video labels alone? And do the gains stem from better directional alignment and layer placement rather than from added parameters?

\subsection{Experimental Setup}
\label{sec:exp_setup}

\paragraph{Datasets and Evaluation Metrics.}
We evaluate on XD-Violence~\citep{wu2020not}, UCF-Crime~\citep{sultani2018real}, and UBnormal~\citep{acsintoae2022ubnormal}, reporting Frame-AP and Frame-AUC for XD-Violence and Frame-AUC for the other two. Training uses binary video-level labels only, with no frame annotations or text descriptions. At test time we map each segment score to its covered frames and compare against the official frame labels. Probes, normalization statistics, and pseudo-labels are fitted on the training splits alone.

\paragraph{Implementation Details.}
Our primary backbone is Qwen3.5-9B~\citep{qwen35blog}, and we also apply RBA to InternVL3.5-8B~\citep{wang2025internvl35} and Gemma-4-12B~\citep{gemma4} to test cross-architecture generality. RBA uses LoRA~\citep{hu2021lora} with rank 16 and scaling factor 32. Guided by the layer-wise diagnostics, we adapt the attention projections in zero-indexed Blocks 20--21 for Qwen3.5-9B, Blocks 25--26 for InternVL3.5-8B, and Blocks 33--34 for Gemma-4-12B. All other weights, including the output embedding and the final RMSNorm scale, stay frozen, leaving about 0.012\% of the backbone trainable. We use AdamW~\citep{loshchilov2017decoupled} with learning rate $1\times10^{-4}$, batch size 32, and three epochs. Training takes under two hours on a single NVIDIA A100 GPU.

\subsection{Main Benchmark Comparison}
\label{sec:main_results}

\begin{table}[!t]
\centering
\small
\setlength{\tabcolsep}{4.5pt}
\renewcommand{\arraystretch}{1.12}
\nobooktabssep
\caption{Quantitative comparison on VAD benchmarks (\%). \cmark{} and \xmark{} denote native support for natural language explanations.}
\label{tab:main_results}
\begin{tabular}{llccccc}
\toprule
\cHB \textbf{Method} & \cHB \textbf{Venue} & \cHB \textbf{Interpretable} 
& \cHB \textbf{UCF AUC} & \cHB \textbf{UB AUC} & \cHB \textbf{XD AUC} & \cHB \textbf{XD AP} \\
\midrule
\multicolumn{7}{l}{\textit{Specialized Detectors (Non-interpretable)}} \\
VadCLIP~\citep{wu2024vadclip} 
& AAAI'24 & \xmark & 88.02 & — & — & \cCO 84.51 \\
$\pi$-VAD~\citep{majhi2025just} 
& CVPR'25 & \xmark & \textbf{90.33} & — & — & \cCO 85.37 \\
RefineVAD~\citep{lee2026refinevad} 
& AAAI'26 & \xmark & 88.92 & — & — & \cCO 88.66 \\
LAVIDA~\citep{dai2026no} 
& CVPR'26 & \xmark & 82.18 & 76.45 & — & \cCO \textbf{90.62} \\
HeadHunt-VAD~\citep{cai2026headhunt} 
& AAAI'26 & \xmark & 87.03 & — & — & \cCO 82.63 \\
PRISM~\citep{chen2026prism} 
& ICML'26 & \xmark & 88.85 & — & 93.42 & \cCO 77.48 \\
\midrule
\multicolumn{7}{l}{\textit{MLLM-based Methods (Interpretable)}} \\
LLaVA-1.5~\citep{liu2024improved} 
& CVPR'24 & \cmark & 72.84 & — & 79.62 & \cCY 50.26 \\
LAVAD~\citep{zanella2024harnessing} 
& CVPR'24 & \cmark & 80.28 & 51.06 & 85.36 & \cCY 62.01 \\
VERA~\citep{ye2025vera} 
& CVPR'25 & \cmark & 86.55 & — & 88.26 & \cCY 70.54 \\
GLM-4.1V-9B~\citep{hong2025glm} 
& - & \cmark & 61.80 & 60.81 & 72.73 & \cCY 52.93 \\
EventVAD~\citep{shao2025eventvad} 
& ACM MM'25 & \cmark & 82.03 & — & 87.51 & \cCY 64.04 \\
Unified-VAD~\citep{lin2026unified} 
& NeurIPS'25 & \cmark & 84.28 & 69.02 & 91.34 & \cCY 68.07 \\
VADTree~\citep{li2026vadtree} 
& NeurIPS'25 & \cmark & 84.74 & 65.80 & 90.55 & \cCY 68.85 \\
PANDA~\citep{yang2026panda} 
& NeurIPS'25 & \cmark & 84.89 & 75.78 & — & \cCY 70.16 \\
Anom-$\pi$~\citep{mo2026learning} 
& ICML'26 & \cmark & 84.46 & \textbf{78.75} & — & \cCY 72.29 \\
LRPO~\citep{leng2026lrpo} 
& ICML'26 & \cmark & 86.59 & 76.24 & — & \cCY 74.09 \\
\cRO \textbf{RBA (Ours)} 
& - & \cmark & \textbf{87.01} & 76.84 & \textbf{94.30} & \cCG \textbf{83.02} \\
\bottomrule
\end{tabular}
\vspace{-4pt}
\end{table}

Table~\ref{tab:main_results} compares RBA with recent VAD methods, which trade off detection accuracy, natural-language output, and weak supervision. Specialized detectors can be accurate yet cannot explain through their decision pathway, while interpretable MLLMs show weaker native readout accuracy or need extra fine-tuning. On the same backbone, RBA lifts the XD-Violence native-readout AP from 72.86\% to 83.02\% and the AUC from 90.37\% to 94.30\%. Among explanation-capable MLLMs it achieves the best result reported here on both XD-Violence and UCF-Crime, exceeding LRPO by 8.93 AP points on XD-Violence. On UBnormal it reaches 76.84\% AUC, above LRPO and PANDA and close to Anom-$\pi$. Aligning the readout direction therefore improves native decisions under weak supervision while retaining language generation.

\subsection{Generalization Across VAD Benchmarks and MLLM Backbones}
\label{sec:generalizability}

\begin{table}[!t]
\centering
\small
\renewcommand{\arraystretch}{1.12}
\nobooktabssep
\caption{Two-factor gap decomposition and alignment gains across benchmarks and backbones, reporting Frame-AP for XD and Frame-AUC for UCF/UB. $P_{\mathrm{full}}$ and $P_{\mathrm{last}}$ are retrospective Fisher ceilings used only for decomposition, whereas $P_{\mathrm{probe}}$ is the weakly supervised attainable ceiling against which the recovered fraction of $\Delta_{\mathrm{dir}}$ is reported in the text. The final column gives the cosine between the Fisher-optimal axis and the native readout axis before and after adaptation.}
\label{tab:gap_decomposition}
\resizebox{\linewidth}{!}{%
\begin{tabular}{lccccccccc}
\toprule
\cHB \textbf{Backbone / Dataset} & \cHB $P_{\mathrm{full}}$ & \cHB $P_{\mathrm{last}}$ & \cHB $P_{\mathrm{probe}}$ & \cHB $P_{\mathrm{read}}$ & \cHB $\Delta_{\mathrm{cap}}$ & \cHB $\Delta_{\mathrm{dir}}$ & \cHB \textbf{Dir. Ratio} & \cHB \textbf{Aligned} & \cHB $\cos(\mathbf{w}_{\mathrm{LDA}}, \mathbf{u})$ \\
\midrule
Qwen3.5-9B / XD & 0.9646 & 0.9588 & 0.8756 & 0.7286 & 0.0058 & 0.2302 & 97.5\% & \cCG \textbf{0.8302} & 0.238 $\rightarrow$ \textbf{0.774} \\
Qwen3.5-9B / UCF & 0.9168 & 0.9120 & 0.8847 & 0.7714 & 0.0048 & 0.1406 & 96.7\% & \cCG \textbf{0.8701} & 0.287 $\rightarrow$ \textbf{0.789} \\
Qwen3.5-9B / UB & 0.8974 & 0.8910 & 0.8523 & 0.7290 & 0.0064 & 0.1620 & 96.2\% & \cCG \textbf{0.7684} & 0.263 $\rightarrow$ \textbf{0.746} \\
InternVL3.5-8B / XD & 0.9251 & 0.9180 & 0.8734 & 0.6493 & 0.0071 & 0.2687 & 97.4\% & \cCG \textbf{0.7842} & 0.164 $\rightarrow$ \textbf{0.728} \\
Gemma-4-12B / XD & 0.9386 & 0.9320 & 0.8938 & 0.6980 & 0.0066 & 0.2340 & 97.3\% & \cCG \textbf{0.8058} & 0.201 $\rightarrow$ \textbf{0.756} \\
\bottomrule
\end{tabular}}
\vspace{-4pt}
\end{table}

Running the three-tier readout chain and the gap decomposition across backbone architectures and video domains yields a gap that is almost entirely directional in every configuration, with $\Delta_{\mathrm{dir}}$ above 96\% of $\Delta_{\mathrm{total}}$ in all five settings and $\Delta_{\mathrm{cap}}$ never exceeding 0.71 points. The cosine similarity between $\mathbf{w}_{\mathrm{LDA}}$ and $\mathbf{u}$ starts below 0.29 everywhere and RBA reliably raises it past 0.72. Measured against the weakly supervised ceiling $P_{\mathrm{probe}}$, the intervention closes 69.1\% of the directional gap on XD-Violence, 87.1\% on UCF-Crime, and 32.0\% on UBnormal, alongside 60.2\% and 55.1\% reductions on InternVL3.5-8B and Gemma-4-12B. Accuracy and axis alignment improve together, which points to directional correction as the main driver of the recovery.

\subsection{Ablation Studies and Mechanistic Verification}
\label{sec:ablation_studies}

\begin{table}[!t]
\centering
\small
\begin{minipage}[t]{0.48\textwidth}
\centering
\setlength{\tabcolsep}{3.0pt}
\renewcommand{\arraystretch}{1.12}
\nobooktabssep
\caption{Adapter placement ablation on XD-Violence (Frame-AP).}
\label{tab:layer_ablation}
\begin{tabular}{lccc}
\toprule
\cHB \textbf{Placement} & \cHB \textbf{Sublayer} & \cHB \textbf{Params} & \cHB \textbf{AP} \\
\midrule
Native Readout & — & 0 & 0.7286 \\
Blocks 5--6 & Post-Attn & $\approx$0.012\% & 0.7314 \\
Blocks 17--18 & Post-Attn & $\approx$0.012\% & 0.7976 \\
\cRO \textbf{Blocks 20--21 (Ours)} & \textbf{Post-Attn} & \textbf{$\approx$0.012\%} & \cCG \textbf{0.8302} \\
Blocks 30--31 & Post-Attn & $\approx$0.012\% & 0.7738 \\
Blocks 20--21 & Post-MLP & $\approx$0.024\% & 0.7415 \\
\bottomrule
\end{tabular}
\end{minipage}
\hfill
\begin{minipage}[t]{0.48\textwidth}
\centering
\setlength{\tabcolsep}{3.2pt}
\renewcommand{\arraystretch}{1.12}
\nobooktabssep
\caption{Decision--explanation consistency under prefix-token intervention (\%).}
\label{tab:faithfulness}
\begin{tabular}{lcccc}
\toprule
\cHB \textbf{Method} & \cHB \textbf{Consist.} & \cHB \textbf{$F^{+}$} & \cHB \textbf{$R^{-}$} & \cHB \textbf{Quality} \\
\midrule
                & 78.2\% & 94.5\% & 86.1\% & \cCO 3.63 \\
\multirow{-2}{*}{Zero-Shot Qwen}  & 79.1\% & 93.8\% & 85.4\% & \cCO 3.57 \\
                & 43.2\% & 71.3\% & 94.4\% & \cCY 2.85 \\
\multirow{-2}{*}{Probe + Caption} & 44.0\% & 70.8\% & 93.9\% & \cCY 2.78 \\
\cRO            & \textbf{91.4\%} & \textbf{96.3\%} & \textbf{36.7\%} & \cCG \textbf{4.12} \\
\cRO \multirow{-2}{*}{\textbf{RBA (Ours)}} & \textbf{92.0\%} & \textbf{95.9\%} & \textbf{35.8\%} & \cCG \textbf{4.09} \\
\bottomrule
\end{tabular}
\vspace{1pt}
\raggedright
{\scriptsize $^\dagger$ First row: GPT-5.4 mini; second row: Gemini 3.6 Flash.}
\end{minipage}
\vspace{-4pt}
\end{table}

\paragraph{Diagnostic-Guided Adapter Placement.}
Table~\ref{tab:layer_ablation} shows that adapters in Blocks 20--21 recover most of the gain with 0.012\% trainable parameters. Blocks 5--6 barely help at 0.7314 AP because the useful features have not yet formed. Blocks 17--18 reach 0.7976 and Blocks 30--31 reach 0.7738, a pattern favoring updates near the start of the transition. Adapting MLP sublayers gives only 0.7415 AP, consistent with attention carrying the positive alignment role.

\paragraph{Decision--Explanation Consistency.}
We test whether explanations agree with decisions using an LLM-based protocol~\citep{liu2023geval} on diagnostic sets of 10,000 clips each. We measure free-generation consistency, compliant faithfulness ($F^{+}$) on correct decisions, and adversarial rationalization ($R^{-}$) when a prefix forces a counterfactual decision. Table~\ref{tab:faithfulness} reports 91.4\% consistency for RBA, 96.3\% $F^{+}$, and 36.7\% $R^{-}$. Probe + Captioning attains only 43.2\% consistency and rationalizes forced counterfactual decisions 94.4\% of the time. A shared generative pathway keeps decisions and explanations consistent, and both judges agree on this ordering.

\subsection{Qualitative Analysis and Geometric Evolution}
\label{sec:qualitative_results}

\begin{figure}[!t]
\centering
\includegraphics[width=\linewidth]{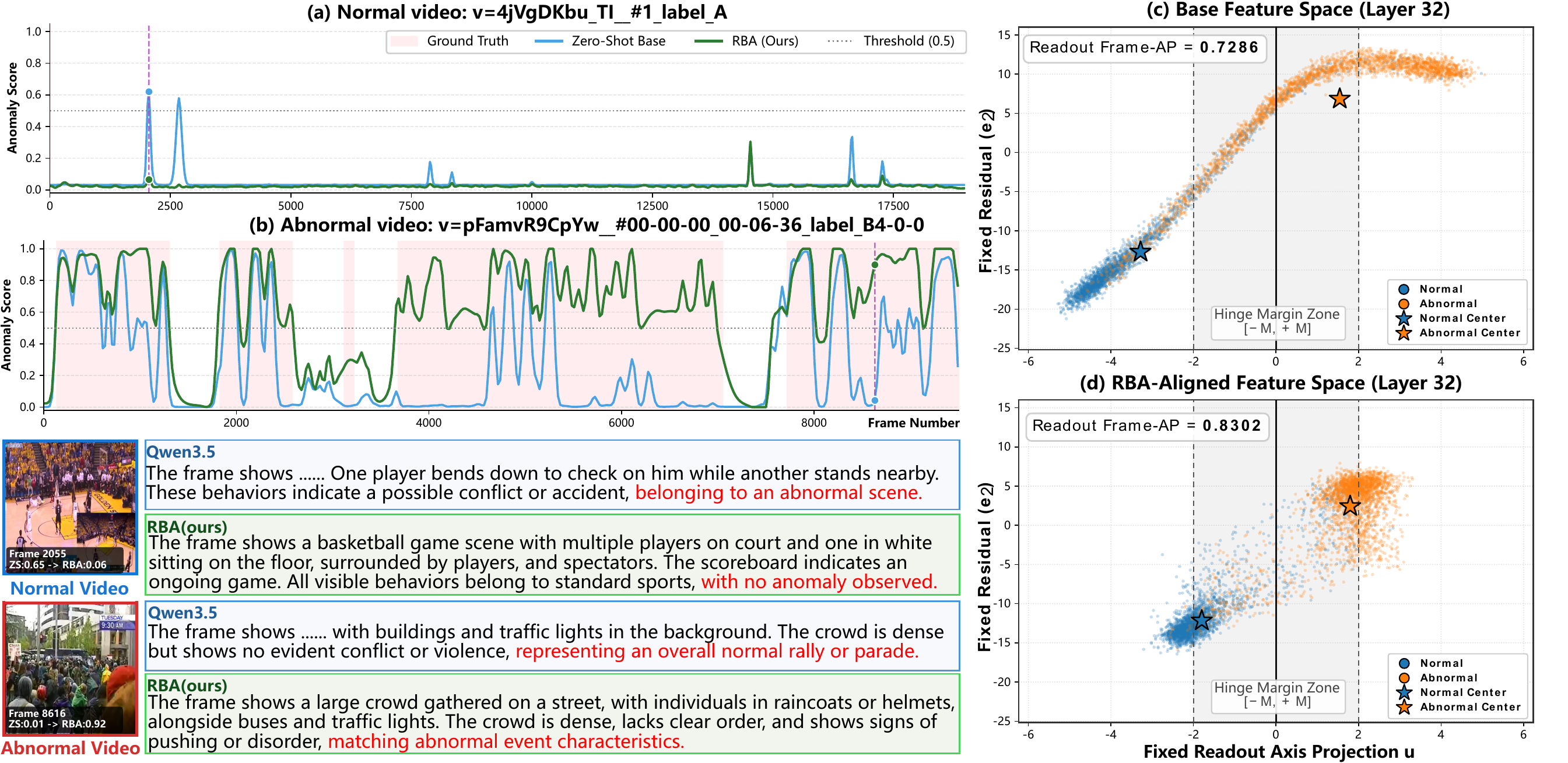}
\caption{Qualitative and geometric analysis. (a)(b) Temporal anomaly scores and corresponding explanations before and after RBA. (c)(d) Final-layer feature distributions along the native readout axis $\mathbf{u}$, reflecting the downstream effect of adapting Blocks 20--21.}
\label{fig:qualitative_and_geometry}
\end{figure}

Figure~\ref{fig:qualitative_and_geometry} shows scores, descriptions, and geometry improving together. In panels (a)(b), the base readout mistakes a player sitting on the floor for an injury and peaks at 0.65 on a normal basketball clip, whereas RBA drops the score to 0.06 and describes standard gameplay. On an abnormal crowd sequence the base model outputs only 0.01, while RBA raises the score to 0.92 and highlights dense pushing and protective helmets. Panels (c)(d) explain this geometrically by projecting Layer~32 representations along the fixed axis $\mathbf{u}$. The two classes heavily overlap inside the hinge band $[-M, +M]$ before alignment, and training the transition-layer adapters pushes representations past the margin boundaries and separates the class centroids cleanly. This redistribution recovers 69.1\% of the directional gap relative to the weakly supervised ceiling $P_{\mathrm{probe}}$ and 44.1\% relative to the retrospective ceiling $P_{\mathrm{last}}$, with the remainder reflecting pseudo-label noise alongside the rank-16 and single-direction constraints. Updating attention projections at transition depth thus redirects features toward the unembedding head without any auxiliary inference classifier.

% -------------------------------------------------------------
% 6. Conclusion
% -------------------------------------------------------------
\section{Conclusion}
We studied the gap between what an MLLM represents and what it says in weakly supervised video anomaly detection. Across three VAD benchmarks and several MLLM backbones, linear probes on hidden states outperform the native normal--abnormal readout. Our single-axis analysis shows that most of the usable gap comes from a mismatch in readout direction, and residual-stream analysis locates the sharp rise in native separation at specific attention sublayers. Based on this result, RBA updates about 0.012\% of the backbone parameters using video labels alone. It improves native anomaly detection while producing explanations through the same generative pathway. We believe representation--behavior alignment may prove equally useful in other multimodal decision tasks.

% -------------------------------------------------------------
% ICLR 声明
% -------------------------------------------------------------
\subsection*{AI Use Statement}
In this work, we used generative AI tools for proofreading and improving text readability. We have not used generative AI tools for scientific idea generation or data simulation. All AI-assisted text was manually reviewed and verified by the authors. The authors take full responsibility for the content of this work.

\subsection*{Ethics Statement}
This work adheres to the ICLR Code of Ethics. The datasets utilized are public benchmark datasets, and this work does not involve direct human subject experiments or known negative social/privacy impacts.

\subsection*{Reproducibility Statement}
To ensure reproducibility, all dataset preprocessing procedures, evaluation protocols, prompt templates, and full hyperparameter configurations are detailed in Appendix~\ref{app:implementation}. The complete source code, trained adapter checkpoints, and offline probing scripts are provided in the supplementary material and will be made publicly available upon release.

% -------------------------------------------------------------
% 致谢 (盲审阶段保持通用)
% -------------------------------------------------------------
%\subsubsection*{Acknowledgments}

% -------------------------------------------------------------
% 参考文献
% -------------------------------------------------------------
\bibliography{iclr2027_conference}
\bibliographystyle{plainnat}

% =============================================================
% Appendix (Complete Corrected Version)
% =============================================================
\clearpage
\appendix
% 引入浮动体控制宏包，防止图片跨小节堆叠
% -------------------------------------------------------------
% Appendix A
% -------------------------------------------------------------
\section{Implementation and Experimental Details}
\label{app:implementation}
\subsection{Backbone Configurations and LoRA Setup}
\label{app:architecture}
Our primary backbone is Qwen3.5-9B, whose language decoder contains 32 blocks indexed from Block~0 to Block~31. Based on the residual-stream diagnostic in Section~\ref{sec:diagnosis}, LoRA modules are inserted into the attention projections of Blocks~20 and~21.
Unless otherwise specified, we use LoRA rank $r=16$, scaling factor $\alpha=32$, and dropout $p_{\mathrm{drop}}=0.05$. The default configuration adapts the query, key, value, output, and architecture-specific auxiliary attention projections. All MLP sublayers, normalization layers, vision encoders, and language unembedding parameters remain frozen.
For a base projection matrix $W_{\mathrm{base}} \in \mathbb{R}^{d_{\mathrm{out}} \times d_{\mathrm{in}}}$, the adapted matrix is
\begin{equation}
W'
=
W_{\mathrm{base}}
+
\frac{\alpha}{r}BA,
\qquad
A\in\mathbb{R}^{r\times d_{\mathrm{in}}},
\quad
B\in\mathbb{R}^{d_{\mathrm{out}}\times r}.
\label{eq:app_lora}
\end{equation}
For cross-backbone diagnostics and evaluations, Gemma~4 and InternVL3.5-8B use the same semantic readout tokens and categorical decision protocol as the primary Qwen3.5-9B experiment. We independently trace the residual stream for each architecture and identify the localized readout transition according to the largest characteristic increase in native-axis separability. The evaluated Gemma~4 and InternVL3.5-8B configurations exhibit their most prominent transitions around Layer~33 and Layer~25, respectively.
For RBA alignment training on InternVL3.5-8B, we follow the same diagnostic-guided placement criterion: adapters are inserted into the identified transition attention block and the immediately subsequent attention block. The exact target projection module names are architecture-specific and are specified in the released configuration files.

\subsection{Probe Training and Weakly-Supervised Pseudo-Labeling}
\label{app:probe_training}
The linear probe $\mathbf{w}_{\mathrm{probe}}\in\mathbb{R}^{d}$ is trained on normalized final-layer hidden states $\tilde{\mathbf{h}}_L$. We use AdamW with an initial learning rate of $1\times 10^{-3}$ and train for 10 epochs.
For video $j$ containing $T_j$ temporal clips, Top-$K$ MIL pooling uses
\begin{equation}
K_j
=
\left\lfloor \frac{T_j}{16}\right\rfloor+1.
\end{equation}
Let
\begin{equation}
s_i
=
\sigma\left(
\mathbf{w}_{\mathrm{probe}}^\top
\tilde{\mathbf{h}}_{L,i}
+b
\right)
\end{equation}
denote the clip-level posterior. The video-level prediction is the mean probability of the $K_j$ highest-scoring clips:
\begin{equation}
\hat{Y}_j
=
\frac{1}{K_j}
\sum_{i\in\operatorname{TopK}(j)}
s_i.
\end{equation}
The probe is trained with video-level binary supervision:
\begin{equation}
\mathcal{L}_{\mathrm{MIL}}
=
\frac{1}{|\mathcal{V}|}
\sum_{j\in\mathcal{V}}
\operatorname{BCE}
\left(Y_j,\hat{Y}_j\right).
\end{equation}
Class balancing is performed strictly at the video-bag level. Clips from the same video are never distributed across different data splits.
After probe training, pseudo-labels are computed only on the training split:
\begin{equation}
\hat{y}_i
=
\begin{cases}
+1, & s_i\geq 0.5,\\
-1, & s_i<0.5.
\end{cases}
\label{eq:app_pseudo_label}
\end{equation}
Frame-level annotations are strictly held out and utilized solely for standard test-set benchmark evaluation and the oracle retrospective analysis presented in Section~\ref{sec:misalignment_diagnosis}. Neither probe training, transition-layer localization, nor RBA adapter optimization utilizes ground-truth frame annotations at any stage.

\paragraph{Leakage-Free Transition Localization via MIL Pseudo-Labels.}
A central operational question is whether the critical intervention depth identified by the retrospective diagnostic in Section~\ref{sec:attribution_and_dynamics} can be reliably recovered without any ground-truth temporal supervision. We demonstrate here that evaluating the residual stream exclusively under training-set MIL pseudo-labels $\hat{y}_i$ consistently pinpoints the exact same transition layer (Blocks~20--21) across all three benchmarks.

Specifically, for each residual-stream sampling point $l$, we evaluate two diagnostic quantities driven solely by pseudo-labels. First, we compute the native abnormal-minus-normal readout score $d_i(l)=\mathbf{u}^\top\tilde{\mathbf{h}}_{l,i}$ and calculate the pseudo-label discriminability:
\begin{equation}
d'_{\mathrm{read}}(l)
=
\frac{
\left|\hat{\mu}_{+}(l)-\hat{\mu}_{-}(l)\right|
}{
\sqrt{\tfrac{1}{2}\left(\hat{\sigma}^{2}_{+}(l)+\hat{\sigma}^{2}_{-}(l)\right)}
},
\label{eq:app_pseudo_dprime}
\end{equation}
where $\hat{\mu}_{\pm}$ and $\hat{\sigma}^{2}_{\pm}$ are the empirical mean and variance of $d_i(l)$ across pseudo-positive and pseudo-negative training clips. Decomposing $d'_{\mathrm{read}}(l)$ into cumulative attention and MLP sublayer contributions precisely identifies the attention update producing the largest discrete increase in native alignment. Second, we record the overall frame-level readout metric (AP or AUC) against $\hat{y}_i$ across depth, which characterizes how native readout quality evolves with depth on the training split alone.

\begin{figure}[!t]
\centering
\includegraphics[width=0.98\linewidth]{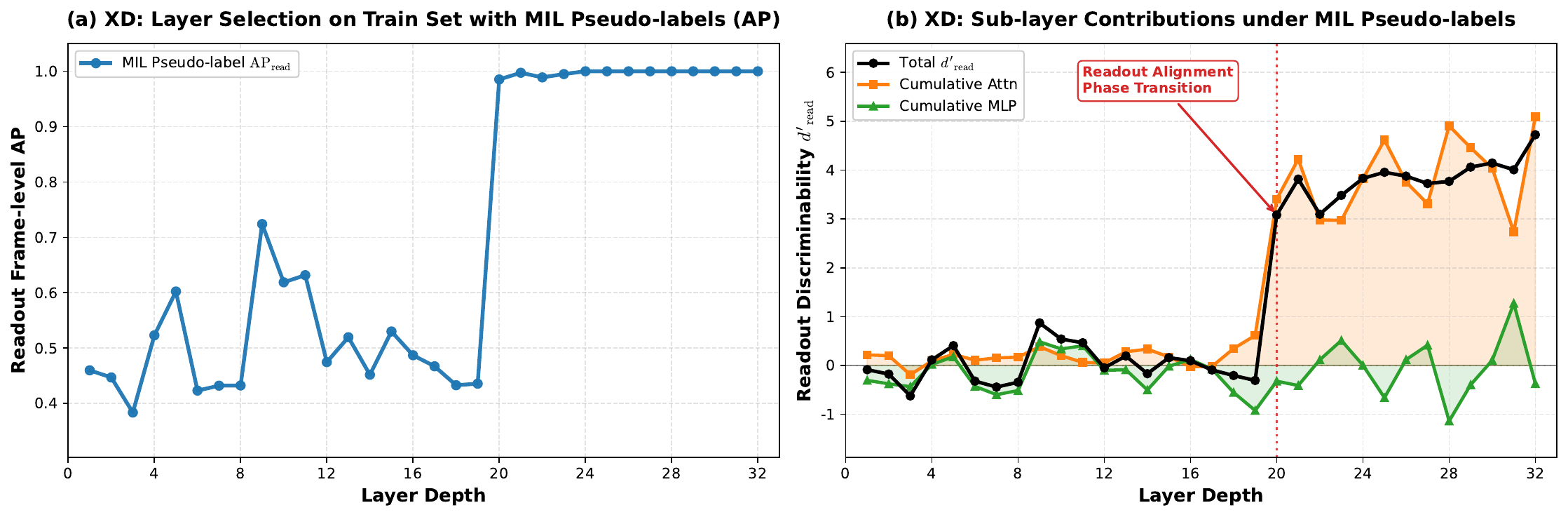}
\caption{Pseudo-label layer diagnosis on the XD-Violence training set (Qwen3.5-9B). (a) Native Frame-AP against pseudo-labels across depth. (b) Pseudo-label discriminability $d'_{\mathrm{read}}$ and cumulative sublayer updates, showing a sharp phase transition at the Block~20 attention sublayer.}
\label{fig:app_layer_sel_xd}
\end{figure}

\begin{figure}[!t]
\centering
\includegraphics[width=0.98\linewidth]{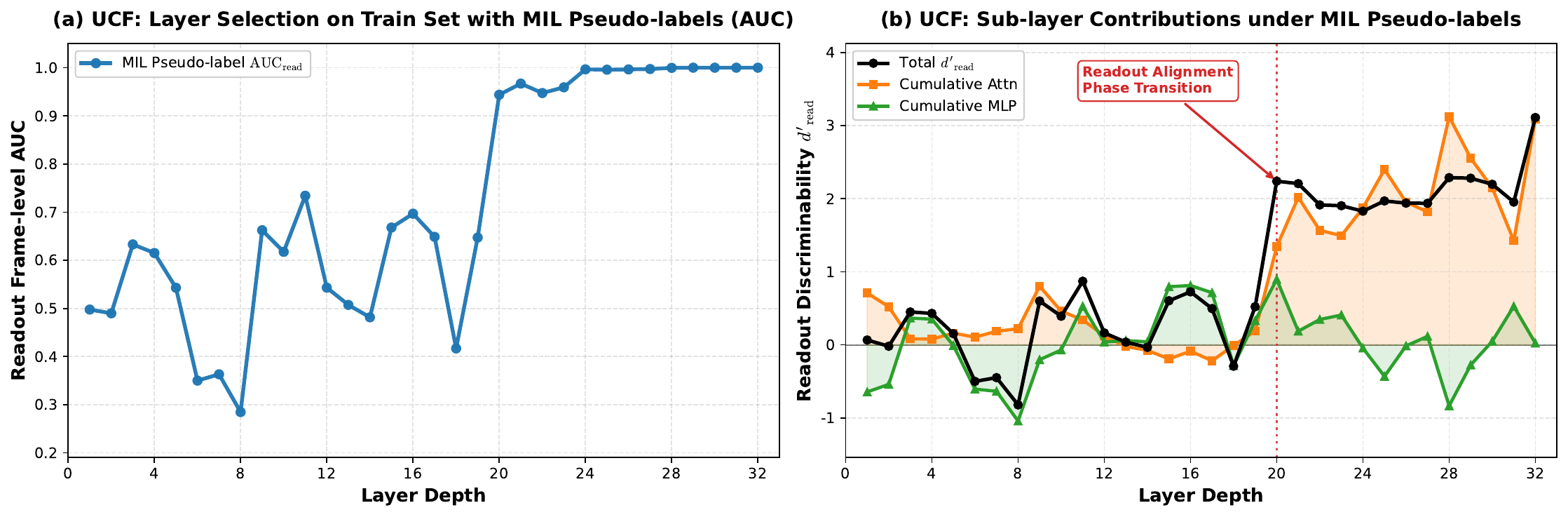}
\caption{Pseudo-label layer diagnosis on the UCF-Crime training set (Qwen3.5-9B). (a) Native Frame-AUC against pseudo-labels across depth. (b) Pseudo-label discriminability $d'_{\mathrm{read}}$, consistently locating the transition at Block~20.}
\label{fig:app_layer_sel_ucf}
\end{figure}

\begin{figure}[!t]
\centering
\includegraphics[width=0.98\linewidth]{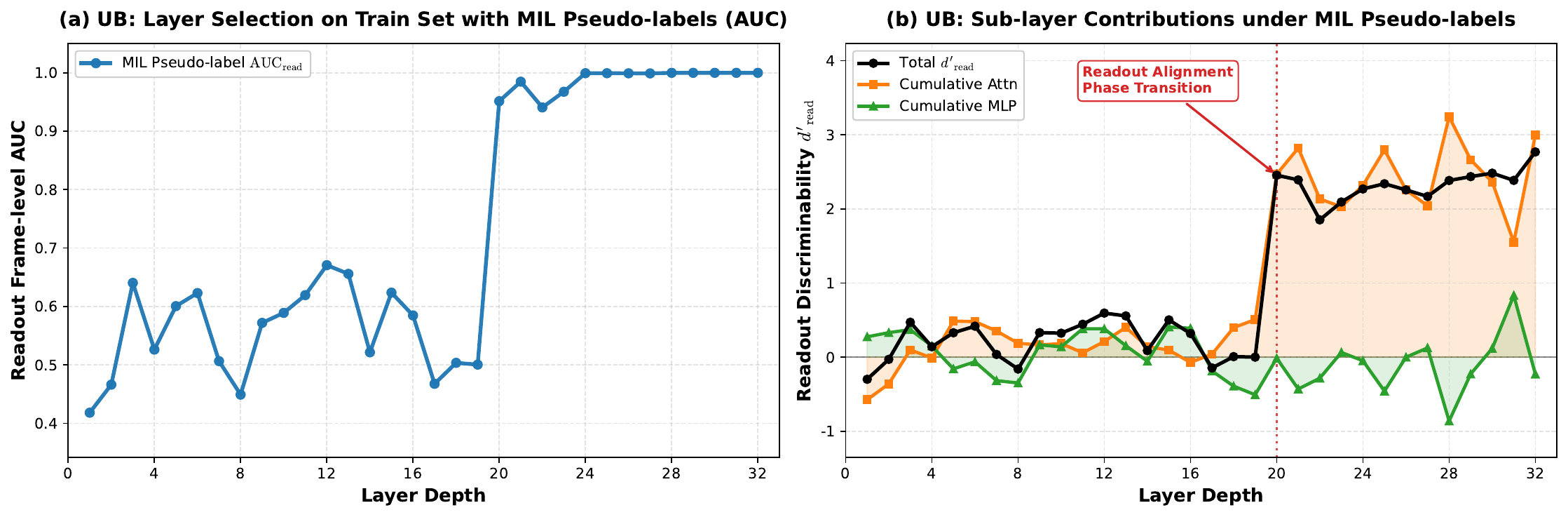}
\caption{Pseudo-label layer diagnosis on the UBnormal training set (Qwen3.5-9B). (a) Native Frame-AUC against pseudo-labels across depth. (b) Pseudo-label discriminability $d'_{\mathrm{read}}$, verifying the identical Block~20 attention transition under synthetic anomaly scenarios.}
\label{fig:app_layer_sel_ub}
\end{figure}

\paragraph{Consistent Transition Identification vs. Superficial Metric Fluctuations.}
Figures~\ref{fig:app_layer_sel_xd}, \ref{fig:app_layer_sel_ucf}, and \ref{fig:app_layer_sel_ub} illustrate the training-set layer-selection trajectories on XD-Violence, UCF-Crime, and UBnormal, respectively. Two decisive observations emerge from these visualizations:

\begin{enumerate}
    \item \textbf{Universal Phase Transition at Blocks 20--21:} As shown in Panel (b) of all three figures, $d'_{\mathrm{read}}$ consistently remains close to zero throughout early-to-mid layers, followed by a dramatic, discontinuous leap at the \textbf{Block~20 attention sublayer} before plateauing. Furthermore, the cumulative decomposition reveals that this sharp emergence is entirely propelled by attention residual updates, while MLP updates produce negative or negligible contributions across all datasets. This proves that the underlying alignment transition is an intrinsic structural landmark of the backbone architecture and can be discovered with 100\% fidelity using only MIL pseudo-labels. Consequently, our policy of injecting adapters into Blocks 20--21 is entirely leakage-free and robustly reproducible across diverse video anomaly domains.

    \item \textbf{Post-Transition Saturation Plateau:} In Panel (a), the readout metric enters an extended, near-flat saturation plateau once the representation passes the transition layer, and the residual variations within this plateau are marginal fluctuations caused by pseudo-label noise and depth-wise representation smoothing. Adapter placement is therefore not driven by where the curve attains its numerical maximum, since deep layers inside the plateau have already undergone heavy representation mixing and offer little headroom for redirecting the decision axis. RBA instead intervenes at the inception of the alignment transition (Blocks~20--21), where attention is actively rotating the decision axis.
\end{enumerate}

In summary, weak-supervision MIL pseudo-labels faithfully preserve both the geometric phase transition and the sublayer opposition discovered in the oracle analysis. The automated detection of Blocks~20--21 across all benchmarks guarantees that RBA operates as a strictly weakly-supervised, leakage-free framework.
\FloatBarrier
\subsection{Margin Alignment Training Protocol}
\label{app:alignment_training}
During alignment fine-tuning, we optimize the hinge-margin objective with margin $M=2.0$:
\begin{equation}
\mathcal{L}_{\mathrm{align}}
=
\frac{1}{B}
\sum_{i=1}^{B}
\max\left(0,M-\hat{y}_i d_i\right),
\end{equation}
where $d_i$ is the abnormal-minus-normal native logit difference.
Optimization uses AdamW with an initial learning rate of $1\times10^{-4}$, cosine learning-rate scheduling, weight decay $0.01$, and a batch size of 32 clips. Training is conducted for three epochs and takes approximately 1.8 hours on one NVIDIA A100 80GB GPU.
All model-selection decisions, including the LoRA rank, dropout, target modules, target blocks, and margin, are made on the validation split. Benchmark test annotations are not used for hyperparameter selection.
\subsection{Standardized Prompt and Native Readout Extraction}
\label{app:prompt}
The following prompt is used consistently for hidden-state extraction, probe training, alignment training, categorical prediction, and open-ended generation:
\begin{quote}
\small
\ttfamily
<|im\_start|>system\\
You are an expert video anomaly detection and reasoning assistant.
<|im\_end|>\\
<|im\_start|>user\\
The following four consecutive video frames record the temporal evolution of a scene. Carefully observe the visual content and determine whether an anomaly or abnormal event occurs:\\[2pt]
[VIDEO\_TOKENS]\\[2pt]
Select the most appropriate option:\\
- Normal\\
- Abnormal\\[2pt]
Formatting requirement: Output only ``Normal'' or ``Abnormal'' on the first line as the final judgment. From the second line onward, briefly describe the visual evidence supporting the judgment.
<|im\_end|>\\
<|im\_start|>assistant\\
Answer:
\end{quote}
The native decision score is the abnormal-minus-normal logit difference at the first answer position:
\begin{equation}
d_i
=
\operatorname{Logit}_i(\mathrm{Abnormal})
-
\operatorname{Logit}_i(\mathrm{Normal}).
\label{eq:app_native_score}
\end{equation}
When either label is tokenized into multiple tokens, we use the teacher-forced sum of conditional token log-probabilities:
\begin{equation}
d_i
=
\log p(\mathrm{Abnormal}\mid V_i,P)
-
\log p(\mathrm{Normal}\mid V_i,P),
\label{eq:app_multitoken_score}
\end{equation}
where $V_i$ denotes the visual input and $P$ denotes the shared prompt.
% -------------------------------------------------------------
% Appendix B (Revised)
% -------------------------------------------------------------
\section{Alignment Design Ablations}
\label{app:alignment_ablations}

This section evaluates the principal design choices of RBA. We first study how video-level supervision is converted into clip-level instances, then examine adapter placement, rank, target modules, dropout, and finally the alignment objective and its margin.

\subsection{MIL Pooling and Instance Mining}
\label{app:mil_ablation}

We compare five strategies for converting video-level supervision into clip-level probe scores:
\begin{enumerate}
    \item \textbf{Mean Pooling:}
    \begin{equation}
    \hat{Y}_j
    =
    \frac{1}{T_j}
    \sum_{t=1}^{T_j}p_{j,t}.
    \end{equation}
    \item \textbf{Max Pooling:}
    \begin{equation}
    \hat{Y}_j
    =
    \max_t p_{j,t}.
    \end{equation}
    \item \textbf{Random-$K$ Pooling:} randomly selects $K_j$ clips and averages their probabilities.
    \item \textbf{Top-$K$ MIL:} averages the $K_j=\lfloor T_j/16\rfloor+1$ highest-scoring clips.
    \item \textbf{Dual-Memory Units:} uses normal and abnormal memory banks $M_n,M_a\in\mathbb{R}^{60\times d}$ following UR-DMU~\citep{zhou2023dual}, before applying Top-$K$ pooling.
\end{enumerate}

\begin{table}[t]
\centering
\small
\setlength{\tabcolsep}{7.5pt}
\caption{Ablation of MIL pooling strategies on downstream RBA performance (\%). Pseudo-label precision and recall are diagnostic quantities computed against training-set temporal annotations; these annotations are not used for optimization.}
\label{tab:app_mil_ablation}
\vspace{2pt}
\begin{tabular}{lcccccc}
\toprule
\textbf{Pooling strategy}
& \textbf{Pseudo-P}
& \textbf{Pseudo-R}
& \textbf{XD AP}
& \textbf{XD AUC}
& \textbf{UCF AUC}
& \textbf{UB AUC}\\
\midrule
Mean Pooling
& 52.41 & 89.67 & 75.14 & 91.23 & 81.38 & 69.82\\
Max Pooling
& 84.16 & 41.29 & 78.43 & 92.47 & 83.94 & 72.18\\
Random-$K$ Pooling
& 56.83 & 74.38 & 76.07 & 91.62 & 82.16 & 70.49\\
\textbf{Top-$K$ MIL}
& \textbf{81.79} & \textbf{79.34}
& \textbf{83.02} & \textbf{94.30}
& \textbf{87.01} & \textbf{76.84}\\
Dual-Memory Units
& 83.27 & 81.08 & 83.46 & 94.51 & 87.23 & 77.12\\
\bottomrule
\end{tabular}
\end{table}

Mean pooling yields high recall but low precision because anomalous segments are diluted by normal background clips. Max pooling has the opposite behavior and is sensitive to isolated high-score clips. Top-$K$ MIL provides a more balanced precision--recall trade-off and substantially improves downstream alignment. Dual-memory mining provides a marginal performance gain (+0.44\% AP) but requires maintaining external bank matrices ($120 \times d$) and prototype distance updates. Top-$K$ MIL introduces no auxiliary parameters or inference overhead, and we adopt it as our default design.

\subsection{Diagnostic-Guided Adapter Placement}
\label{app:placement_ablation}

We evaluate adapter placement at different depths while keeping the attention parameter budget strictly controlled.

\begin{table}[t]
\centering
\small
\setlength{\tabcolsep}{5pt}
\caption{Adapter placement ablation on XD-Violence. All attention placements use approximately the same trainable parameter budget.}
\label{tab:app_placement_ablation}
\begin{tabular}{lccc}
\toprule
\textbf{Placement}
& \textbf{Sublayer}
& \textbf{Params}
& \textbf{Frame-AP}\\
\midrule
Native Readout
& --
& 0
& 0.7286\\
Blocks 5--6
& Post-Attention
& $\approx 0.012\%$
& 0.7314\\
Blocks 17--18
& Post-Attention
& $\approx 0.012\%$
& 0.7976\\
\textbf{Blocks 20--21}
& \textbf{Post-Attention}
& $\mathbf{\approx 0.012\%}$
& \textbf{0.8302}\\
Blocks 30--31
& Post-Attention
& $\approx 0.012\%$
& 0.7738\\
Blocks 20--21
& Post-MLP
& $\approx 0.024\%$
& 0.7415\\
\bottomrule
\end{tabular}
\end{table}

The strong dependence on placement indicates that the improvement is not explained by parameter count alone. Adapters at Blocks 20--21 are located at the empirically identified readout transition and achieve the largest gain. Notably, placing adapters on the post-MLP sublayers yields an inferior 0.7415 AP despite consuming twice the parameter budget ($\approx 0.024\%$). This confirms that the positive native-axis alignment observed in the residual stream is driven specifically by attention sublayers.

\subsection{LoRA Rank and Target Module Ablations}
\label{app:lora_ablation}

\begin{table}[t]
\centering
\small
\setlength{\tabcolsep}{6pt}
\caption{LoRA rank ablation in Blocks 20--21 with $\alpha/r=2.0$.}
\label{tab:app_lora_rank}
\begin{tabular}{cccccc}
\toprule
\textbf{Rank $r$}
& \textbf{Params}
& \textbf{XD AP}
& \textbf{XD AUC}
& \textbf{UCF AUC}
& \textbf{UB AUC}\\
\midrule
1  & $\approx 0.001\%$ & 77.83 & 92.14 & 82.47 & 73.19\\
4  & $\approx 0.003\%$ & 80.46 & 93.18 & 84.82 & 74.93\\
8  & $\approx 0.006\%$ & 82.17 & 93.84 & 86.29 & 76.15\\
\textbf{16}
& $\mathbf{\approx 0.012\%}$
& \textbf{83.02}
& \textbf{94.30}
& \textbf{87.01}
& \textbf{76.84}\\
32 & $\approx 0.024\%$ & 83.09 & 94.36 & 87.04 & 76.81\\
64 & $\approx 0.048\%$ & 82.93 & 94.25 & 86.97 & 76.68\\
\bottomrule
\end{tabular}
\end{table}

Rank-one adaptation already improves XD-Violence Frame-AP from 72.86\% to 77.83\%, showing that a low-dimensional intervention can recover part of the native-readout gap. Performance largely saturates at $r=16$; larger ranks provide negligible gains.

\begin{table}[t]
\centering
\small
\setlength{\tabcolsep}{6pt}
\caption{Ablation of target attention projections in Blocks 20--21 using $r=16$ and $\alpha=32$.}
\label{tab:app_lora_modules}
\begin{tabular}{lcccc}
\toprule
\textbf{Target projections}
& \textbf{Params}
& \textbf{XD AP}
& \textbf{UCF AUC}
& \textbf{UB AUC}\\
\midrule
Query, key, and value ($q,k,v$)
& $\approx 0.006\%$ & 80.67 & 85.14 & 75.32\\
Attention output ($o$)
& $\approx 0.002\%$ & 79.83 & 84.67 & 74.89\\
Standard attention ($q,k,v,o$)
& $\approx 0.008\%$ & 82.14 & 86.38 & 76.23\\
\textbf{All attention projections}
& $\mathbf{\approx 0.012\%}$
& \textbf{83.02}
& \textbf{87.01}
& \textbf{76.84}\\
\bottomrule
\end{tabular}
\end{table}

Adapting both the input and output sides of attention performs better than adapting either side alone. Including all projection matrices within the attention block (covering standard $q, k, v, o$ along with the projection heads) provides a further gain of 0.88\% AP, confirming that comprehensive attention transformation best guides directional rotation.

\subsection{Dropout and Training Stability}
\label{app:dropout_ablation}

We evaluate $p_{\mathrm{drop}}\in\{0,0.05,0.1,0.2\}$ across five random seeds. The default setting $p_{\mathrm{drop}}=0.05$ obtains $83.02\pm0.17$ AP on XD-Violence, compared with $82.46\pm0.24$ for $p_{\mathrm{drop}}=0$ and $82.13\pm0.29$ for $p_{\mathrm{drop}}=0.2$. These results suggest that mild dropout improves robustness to noisy pseudo-labels without materially reducing native-axis alignment.

\subsection{Margin Sensitivity}
\label{app:margin_ablation}

We evaluate the hinge margin $M$ while retaining the default LoRA rank, target layers, optimizer, and pseudo-labeling procedure.

\begin{table}[t]
\centering
\small
\setlength{\tabcolsep}{6pt}
\caption{Sensitivity of the hinge margin $M$ on XD-Violence.}
\label{tab:app_margin}
\begin{tabular}{lcccccc}
\toprule
\textbf{Margin $M$}
& 0.5 & 1.0 & 1.5 & \textbf{2.0} & 2.5 & 3.0\\
\midrule
Frame-AP
& 78.94 & 81.23 & 82.68 & \textbf{83.02} & 82.89 & 81.95\\
Frame-AUC
& 92.81 & 93.74 & 94.18 & \textbf{94.30} & 94.22 & 93.80\\
\bottomrule
\end{tabular}
\end{table}

Performance is stable for $M\in[1.5,2.5]$ and peaks at $M=2.0$. Small margins provide insufficient separation, whereas overly large margins continue to optimize hard or noisy pseudo-labeled clips after the useful directional correction has largely been achieved.

\subsection{Comparison of Alignment Objectives}
\label{app:loss_ablation}

To distinguish directional alignment from probability calibration, we compare the proposed hinge loss with BCE, KL, MSE, and joint decision--explanation training. Let
\begin{equation}
z_i=\frac{\hat{y}_i+1}{2}\in\{0,1\}.
\end{equation}
The compared objectives are
\begin{align}
\mathcal{L}_{\mathrm{BCE}}
&=
-\frac{1}{B}\sum_{i=1}^{B}
\left[
z_i\log\sigma(d_i)
+
(1-z_i)\log(1-\sigma(d_i))
\right],\\
\mathcal{L}_{\mathrm{KL}}
&=
\frac{1}{B}\sum_{i=1}^{B}
D_{\mathrm{KL}}
\left(
\operatorname{Bern}(s_i)
\,\Vert\,
\operatorname{Bern}(\sigma(d_i))
\right),\\
\mathcal{L}_{\mathrm{MSE}}
&=
\frac{1}{B}\sum_{i=1}^{B}
\left(d_i-\hat{y}_iM\right)^2,\\
\mathcal{L}_{\mathrm{joint}}
&=
\mathcal{L}_{\mathrm{align}}
+
\lambda\mathcal{L}_{\mathrm{explain}}.
\end{align}

\begin{table}[t]
\centering
\small
\setlength{\tabcolsep}{5pt}
\caption{Comparison of alignment objectives on XD-Violence. The metric $\cos(\mathbf{w}'_{\mathrm{probe}}, \mathbf{u})$ evaluates directional alignment between the adapted probe direction and the fixed native axis.}
\label{tab:app_loss_ablation}
\begin{tabular}{lccc}
\toprule
\textbf{Training objective}
& \textbf{Frame-AP}
& \textbf{Frame-AUC}
& $\cos(\mathbf{w}'_{\mathrm{probe}},\mathbf{u})$\\
\midrule
Zero-Shot Base Readout
& 72.86 & 90.37 & 0.238\\
$\mathcal{L}_{\mathrm{BCE}}$ (hard labels)
& 79.82 & 93.12 & 0.684\\
$\mathcal{L}_{\mathrm{KL}}$ (soft distribution)
& 78.45 & 92.68 & 0.641\\
$\mathcal{L}_{\mathrm{MSE}}$ (logit regression)
& 77.19 & 91.85 & 0.612\\
$\mathcal{L}_{\mathrm{align}}+
 \mathcal{L}_{\mathrm{explain}}$
& 72.79 & 90.41 & 0.298\\
\textbf{$\mathcal{L}_{\mathrm{align}}$}
& \textbf{83.02} & \textbf{94.30} & \textbf{0.781}\\
\bottomrule
\end{tabular}
\end{table}

The hinge objective achieves the highest native detection accuracy along with the largest improvement in geometric alignment. Objectives like BCE and KL continue enforcing probability calibration even after representations cross the decision boundary, which misallocates the limited rank-16 capacity. Such probability fitting becomes counterproductive when dealing with potentially noisy pseudo-labels.

Introducing an auxiliary explanation loss alongside the margin term degrades both AP and AUC back to baseline levels, keeping the cosine similarity at a low value of 0.298. Autoregressive language modeling gradients diffuse the low-rank updates across numerous token positions, thereby diluting the targeted geometric correction. These comparisons demonstrate that placing adapters at the transition layers works best under a focused native-axis margin loss, which directly enhances boundary separation without requiring probability calibration.
% -------------------------------------------------------------
% Appendix C
% -------------------------------------------------------------
\section{Mechanistic Interpretation of the Readout Transition}
\label{app:mechanism}
\subsection{Potential Sources of Readout Misalignment}
\label{app:misalignment_sources}
Several properties of general-purpose MLLMs may contribute to the representation--behavior gap:
\begin{enumerate}
    \item \textbf{Multi-task pretraining.}
    Hidden states must support diverse visual and linguistic tasks rather than a single VAD decision boundary.
    \item \textbf{Instruction post-training.}
    Instruction tuning optimizes broadly useful language behavior and does not directly optimize temporal anomaly ranking.
    \item \textbf{Task-independent lexical readout.}
    The native direction
    \begin{equation}
    \mathbf{u}
    =
    \frac{
    \mathbf{W}_U[\mathrm{Abnormal}]
    -
    \mathbf{W}_U[\mathrm{Normal}]
    }{
    \left\|
    \mathbf{W}_U[\mathrm{Abnormal}]
    -
    \mathbf{W}_U[\mathrm{Normal}]
    \right\|_2
    }
    \end{equation}
    is inherited from pretraining rather than fitted to VAD data.
\end{enumerate}
These factors provide plausible explanations for the measured angular difference between the weakly supervised probe and the native readout. They are not independently verified causal mechanisms.
\subsection{Empirical Progression Across Depth}
\label{app:depth_progression}
Probe performance rises rapidly and reaches approximately $0.85$ AP around Block~15, indicating that linearly accessible anomaly information is already present before the final decoder blocks.
On Qwen3.5-9B with XD-Violence, native-axis separability increases sharply at the attention sublayer of Block~20:
\begin{equation}
d'_{\mathrm{read}}: 0.474\longrightarrow 1.545.
\end{equation}
We refer to this empirically observed region as the \emph{readout transition region}. The term describes the measured behavior of this model--dataset configuration and does not imply a universal mathematical phase transition.
After the transition, native-axis separability increases more slowly and eventually plateaus. Together with the adapter placement and objective ablations, this suggests that anomaly-discriminative features form early, while their projection into the language generation space is refined later.
\subsection{Robustness Across Readout Tokens}
\label{app:token_robustness}
The primary diagnostic in Figure~\ref{fig:phase_transition} employs the default single-token verbalizers (\texttt{Normal} and \texttt{Abnormal}). To examine whether the observed depth-wise dynamics are sensitive to lexical realization, we keep the primary Qwen3.5-9B backbone, video inputs, prompt structure, extraction position, and evaluation protocol strictly fixed, and evaluate two alternative readout formulations.
For any pair of single-token verbalizers $(\tau_{\mathrm{abn}},\tau_{\mathrm{nor}})$, the native decision axis is defined as
\begin{equation}
\mathbf{u}_{\tau}
=
\frac{
\mathbf{W}_{U}[\tau_{\mathrm{abn}}]
-
\mathbf{W}_{U}[\tau_{\mathrm{nor}}]
}{
\left\|
\mathbf{W}_{U}[\tau_{\mathrm{abn}}]
-
\mathbf{W}_{U}[\tau_{\mathrm{nor}}]
\right\|_2
}.
\label{eq:app_token_axis}
\end{equation}
When verbalizers comprise multiple sub-tokens, we use the teacher-forced log-probability difference in Equation~\ref{eq:app_multitoken_score}.
\begin{figure}[!t]
\centering
\includegraphics[width=0.98\linewidth]{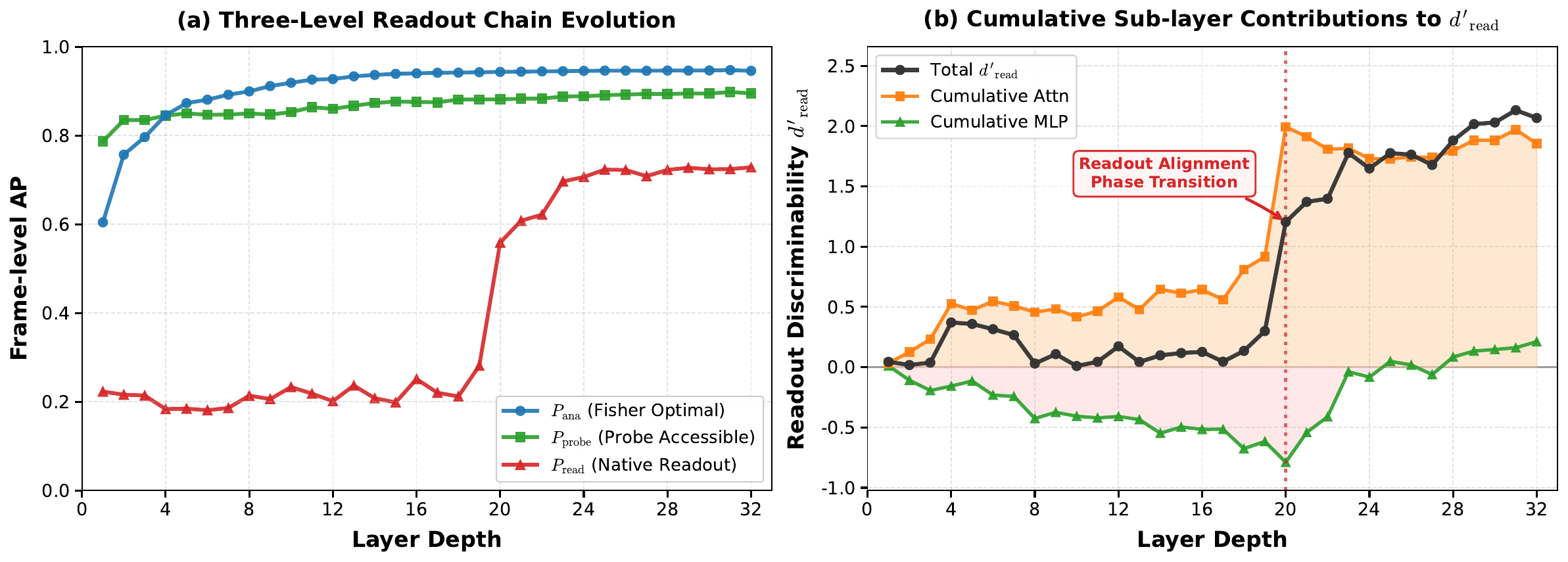}
\caption{Layer-wise residual-stream diagnostic on Qwen3.5-9B using alternative word-level readout tokens (\texttt{Yes}/\texttt{No}) on XD-Violence. Left: three-tier readout trajectories across network depth. Right: native discriminability $d'_{\mathrm{read}}$ and cumulative sublayer contributions.}
\label{fig:app_diag_word}
\end{figure}
\begin{figure}[!t]
\centering
\includegraphics[width=0.98\linewidth]{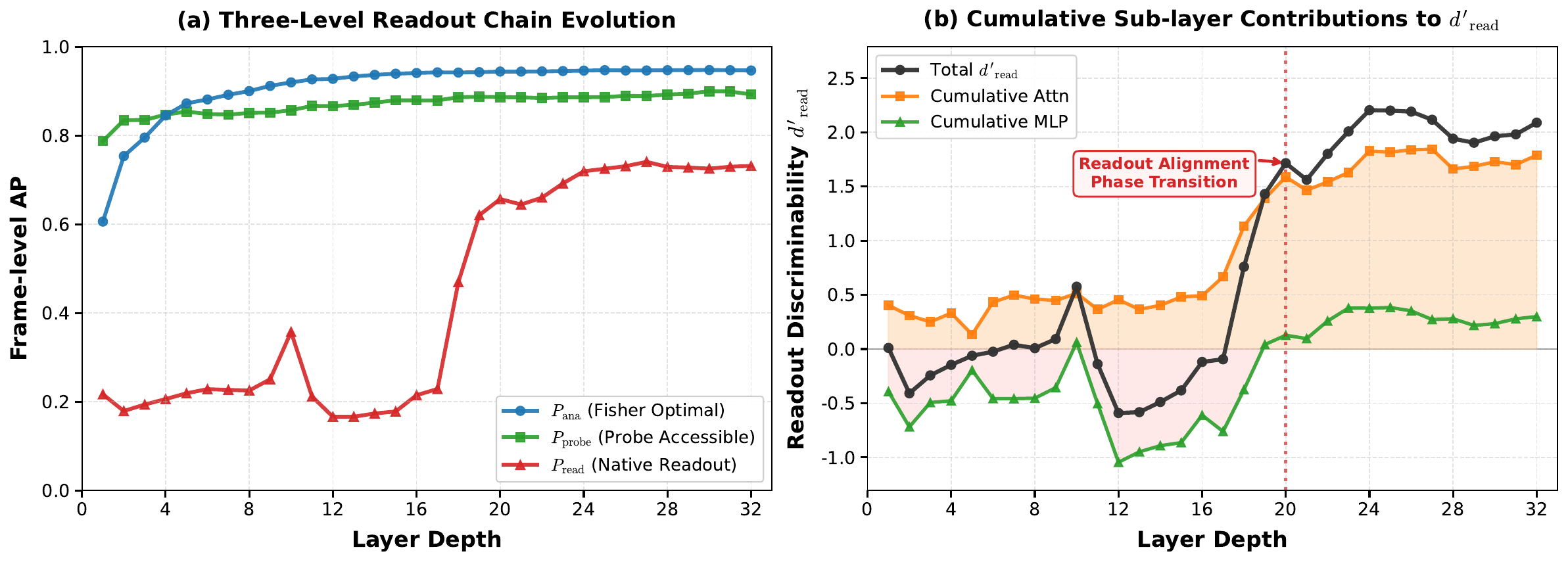}
\caption{Layer-wise residual-stream diagnostic on Qwen3.5-9B using \texttt{Safe}/\texttt{Risky} readout tokens on XD-Violence. Left: three-tier readout trajectories across network depth. Right: native discriminability $d'_{\mathrm{read}}$ and cumulative sublayer contributions.}
\label{fig:app_diag_saferisk}
\end{figure}
As shown in Figures~\ref{fig:app_diag_word} and~\ref{fig:app_diag_saferisk}, changing the readout tokens shifts the absolute numerical baseline because different token embeddings occupy distinct coordinates in the unembedding space. Crucially, both token variants preserve the central diagnostic traits:
\begin{enumerate}
    \item \textbf{Early representation maturity:} The weakly-supervised linear probe ($P_{\mathrm{probe}}$) ascends early in the network, confirming that anomaly features are formed well before native readout emergence.
    \item \textbf{Localized transition:} Native readout discriminability undergoes a sharp, localized increase in the mid-to-late attention layers rather than accumulating linearly from the input.
    \item \textbf{Sublayer dynamics:} Positive alignment gains remain driven predominantly by attention residual updates, while MLP sublayers provide smaller or opposing contributions.
\end{enumerate}
These findings demonstrate that the representation--behavior gap is an intrinsic property of the model's depth-wise processing rather than an artifact of a specific lexical choice.
\FloatBarrier
\subsection{Robustness Across MLLM Backbones}
\label{app:backbone_robustness}
To assess cross-architecture generality, we evaluate the diagnostic pipeline on Gemma~4 and InternVL3.5-8B while strictly maintaining the main-experiment readout tokens, categorical protocol, and XD-Violence dataset split.
\begin{figure}[!t]
\centering
\includegraphics[width=0.98\linewidth]{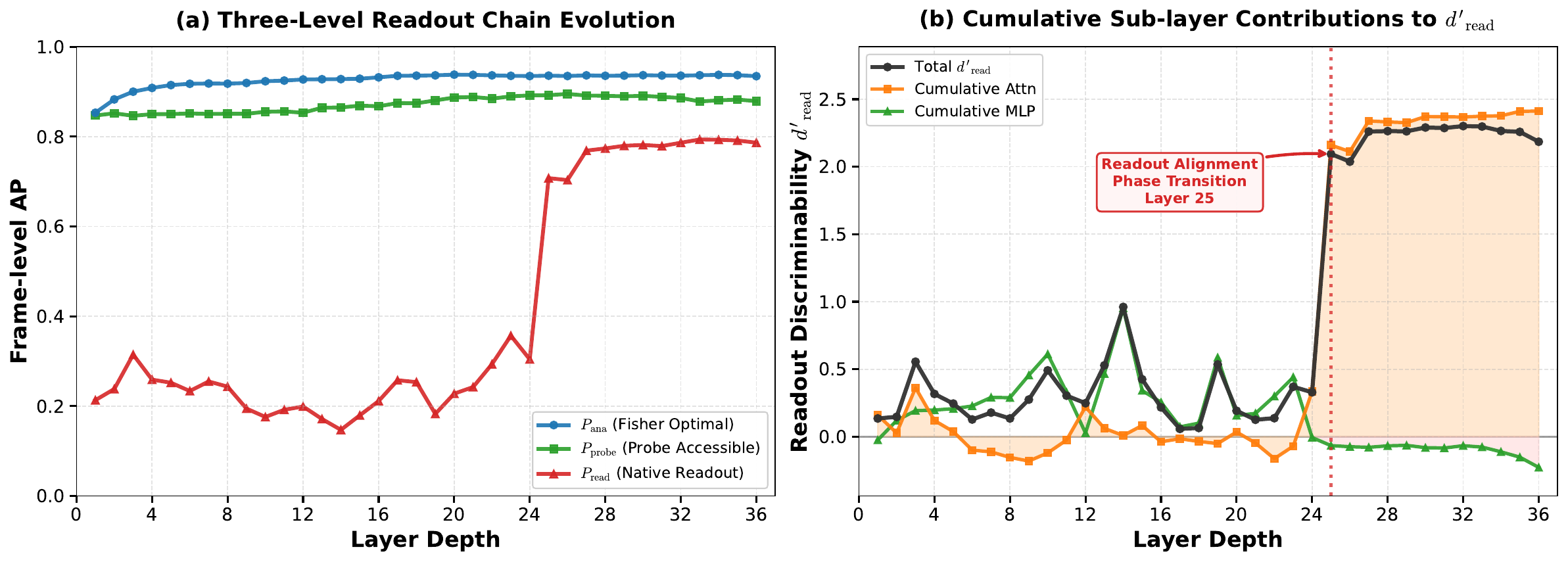}
\caption{Layer-wise residual-stream diagnostic on InternVL3.5-8B on XD-Violence using the main-experiment readout tokens. Left: three-tier readout trajectories across network depth. Right: native discriminability $d'_{\mathrm{read}}$ and cumulative sublayer contributions.}
\label{fig:app_diag_internvl}
\end{figure}
\begin{figure}[!t]
\centering
\includegraphics[width=0.98\linewidth]{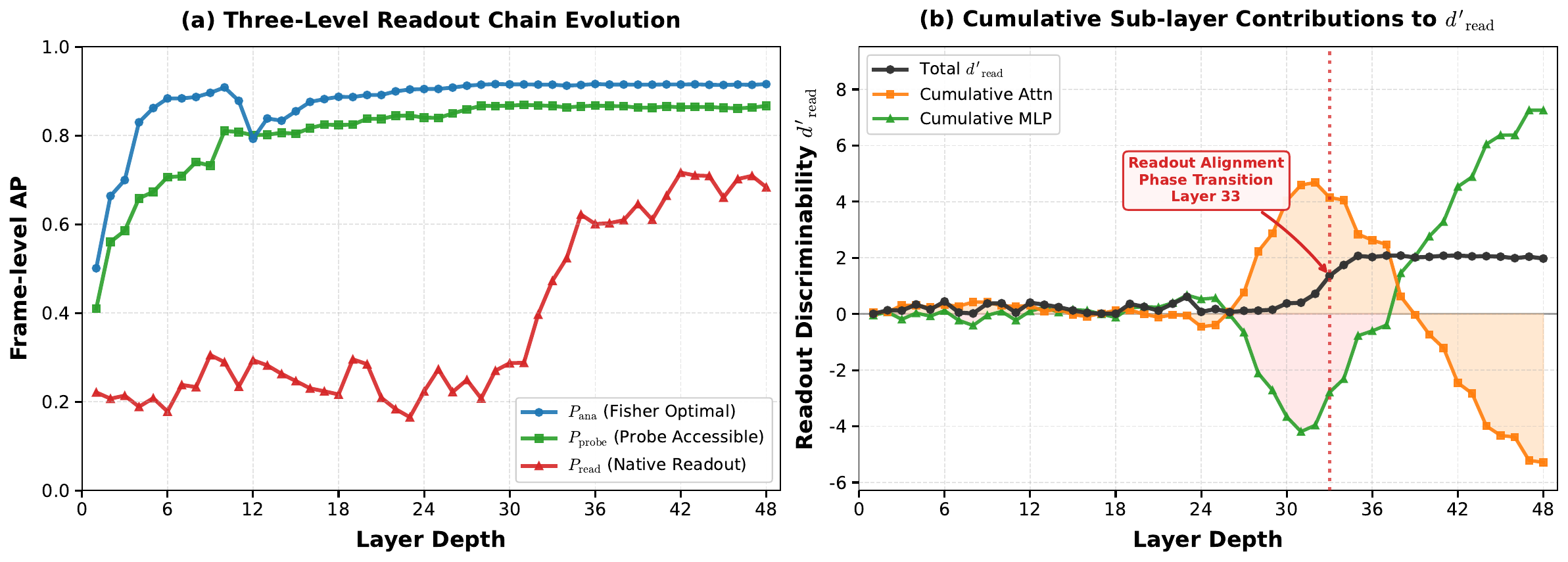}
\caption{Layer-wise residual-stream diagnostic on Gemma~4 on XD-Violence using the main-experiment readout tokens. Left: three-tier readout trajectories across network depth. Right: native discriminability $d'_{\mathrm{read}}$ and cumulative sublayer contributions.}
\label{fig:app_diag_gemma}
\end{figure}
Figures~\ref{fig:app_diag_gemma} and~\ref{fig:app_diag_internvl} confirm that the structural separation between representation accessibility and native readout alignment generalizes across diverse MLLM architectures. In both models, the probe reaches near-peak discriminative performance early, whereas the native readout remains unaligned until a localized transition in the deeper layers.
Due to architectural variations in depth, attention topology, and pretraining distributions, the transition occurs at backbone-dependent depths: around Layer~33 in Gemma~4 and Layer~25 in InternVL3.5-8B. Across both architectures, attention residual updates remain the primary source of positive alignment gains. This validates our methodological premise: rather than hardcoding a fixed layer index, RBA applies a diagnostic-guided principle that localizes the critical transition layers specific to each backbone prior to adapter placement.
\FloatBarrier
\subsection{Objective-Based Evidence for Transition-Layer Specialization}
\label{app:transition_specialization}
The loss comparison in Table~\ref{tab:app_loss_ablation} provides functional evidence regarding transition-layer specialization. If these layers merely enhanced generic anomaly separability without directional preference, distinct objectives would produce comparable improvements under the same parameter constraints. Instead, the native-axis hinge loss yields the strongest detection metrics and maximizes the alignment value $\cos(\mathbf{w}'_{\mathrm{probe}},\mathbf{u})$. The inferior outcomes of BCE and KL suggest that matching probe probabilities is unnecessary for decision boundary correction. Furthermore, joint explanation training damages detection accuracy because dispersing the low-rank capacity over multi-token sequences compromises the primary rotation objective. These results indicate that the identified transition layers are well suited for an efficient representation-to-readout alignment rather than broad semantic fine-tuning.
\label{app:attention_mlp}
Residual-update decomposition shows that attention sublayers contribute positive gains to native-axis separability, whereas MLP sublayers introduce negative interference in the examined configuration. The post-attention placement result in Table~\ref{tab:app_placement_ablation}, together with the inferior post-MLP result, provides a parameter-controlled empirical check of this observation.
This pattern motivates freezing MLP sublayers during RBA training. The purpose is not to assume that MLPs are uninformative, but to avoid using the limited low-rank adaptation budget to modify components whose measured net contribution is not aligned with the desired native-axis correction.
\subsection{Null-Space Diagnostic}
\label{app:null_space}
We remove the normalized LDA direction from the final hidden states:
\begin{equation}
\mathbf{h}^{\perp}
=
\left(
\mathbf{I}
-
\hat{\mathbf{w}}_{\mathrm{LDA}}
\hat{\mathbf{w}}_{\mathrm{LDA}}^\top
\right)
\tilde{\mathbf{h}}_L.
\end{equation}
A new probe trained and evaluated in the projected space obtains $0.5237$ AP, compared with $0.8756$ before projection. This result shows that the removed direction contains a large fraction of the linearly accessible discriminative signal. It does not rule out additional nonlinear or distributed anomaly information in the remaining subspace.
\subsection{Implications for RBA}
\label{app:mechanism_implications}
The diagnostics motivate three design choices:
\begin{itemize}
    \item \textbf{Targeted placement.}
    Adapters are placed near the empirically identified readout transition.
    \item \textbf{Attention-only adaptation.}
    Attention projections are adapted while MLPs and the unembedding head remain frozen.
    \item \textbf{Native-axis optimization.}
    The hinge objective directly increases class separation along the fixed native decision axis without requiring probability matching.
\end{itemize}
% -------------------------------------------------------------
% Appendix D
% -------------------------------------------------------------
\section{Decision--Explanation Evaluation Protocol}
\label{app:faithfulness}
\subsection{Diagnostic Sets}
\label{app:faithfulness_sets}
We evaluate decision--explanation behavior on fixed standardized sets constructed from the official test splits of XD-Violence, UCF-Crime, and UBnormal.
The compliant faithfulness set $\mathcal{D}^{+}$ contains $N^{+}=10{,}000$ clips, balanced between normal and abnormal examples, for which the model's native decision is correct. The adversarial rationalization set $\mathcal{D}^{-}$ contains $N^{-}=10{,}000$ clips subjected to counterfactual decision forcing.
Each clip contains four temporally ordered frames, a binary ground-truth label $y_i$, and a human-verified reference description $\mathcal{C}^{\mathrm{ref}}_i$. Sampling is stratified across datasets and classes using seed 42.
\subsection{Compared Generation Pipelines}
\label{app:faithfulness_pipelines}
We compare:
\begin{itemize}
    \item \textbf{Zero-Shot Qwen:} the frozen backbone directly generates the categorical decision and explanation.
    \item \textbf{Probe + Caption:} the external probe supplies the categorical decision, while the frozen backbone independently generates the explanation.
    \item \textbf{RBA:} the aligned model generates both the categorical decision and explanation through one autoregressive pathway.
\end{itemize}
Greedy decoding is used for all methods. The categorical decision is parsed only from the first generated line. Invalid or unparsable first lines are counted as incorrect decisions and inconsistent outputs.
For the adversarial benchmark, the first-line decision is replaced by the opposite label while keeping the visual input fixed. The model then generates the explanation conditioned on the forced prefix.
\subsection{LLM-as-a-Judge Evaluation}
\label{app:faithfulness_judge}
We use two independent judges with temperature zero. Each judge returns a consistency label $c_i^m\in\{0,1\}$ and an evidence-faithfulness grade $g_i^m\in\{\mathrm{A},\mathrm{B},\mathrm{C},\mathrm{D}\}$.
The judge receives the ground-truth label, reference description, model decision, and model explanation, but not the method name.
\begin{quote}
\small
\ttfamily
You are an impartial evaluator of video anomaly decisions and explanations. Evaluate the model output using the supplied reference.\\[2pt]
Ground-truth label: \{ground\_truth\_label\}\\
Reference description: \{reference\_description\}\\
Model decision: \{model\_decision\}\\
Model explanation: \{model\_explanation\}\\[2pt]
Return exactly two fields:\\
Consistency: YES or NO\\
Faithfulness: A, B, C, or D
\end{quote}
Consistency evaluates whether the explanation supports the model's own decision, independently of whether that decision is correct. Faithfulness evaluates the explanation against the reference evidence.
\subsection{Metrics}
\label{app:faithfulness_metrics}
We map the faithfulness grade to
\begin{equation}
\mathcal{Q}(g)
=
\begin{cases}
5, & g=\mathrm{A},\\
4, & g=\mathrm{B},\\
2, & g=\mathrm{C},\\
0, & g=\mathrm{D}.
\end{cases}
\end{equation}
Decision--explanation consistency is
\begin{equation}
\operatorname{Consist.}^{m}
=
\frac{1}{N}
\sum_{i=1}^{N}c_i^m
\times 100\%.
\end{equation}
Compliant faithfulness is
\begin{equation}
F_m^{+}
=
\frac{1}{N^{+}}
\sum_{i\in\mathcal{D}^{+}}
\frac{\mathcal{Q}(g_i^m)}{5}
\times 100\%.
\end{equation}
The adversarial rationalization rate is
\begin{equation}
R_m^{-}
=
\frac{1}{N^{-}}
\sum_{i\in\mathcal{D}^{-}}
\mathbb{I}
\left[
c_i^m=1
\land
g_i^m\in\{\mathrm{C},\mathrm{D}\}
\right]
\times 100\%.
\end{equation}
Lower $R^{-}$ indicates greater sensitivity to the visual evidence under counterfactual decision forcing. Overall explanation quality is
\begin{equation}
\operatorname{Quality}^{m}
=
\frac{1}{N}
\sum_{i=1}^{N}
\mathcal{Q}(g_i^m).
\end{equation}
\begin{table}[t]
\centering
\small
\setlength{\tabcolsep}{5pt}
\caption{Decision--explanation evaluation on fixed diagnostic sets. Consistency and $F^{+}$ are higher-is-better, while $R^{-}$ is lower-is-better.}
\label{tab:app_faithfulness}
\begin{tabular}{llcccc}
\toprule
\textbf{Method}
& \textbf{Judge}
& \textbf{Consist.}
& $\mathbf{F^{+}}$
& $\mathbf{R^{-}}$
& \textbf{Quality}\\
\midrule
Zero-Shot Qwen
& GPT-5.4 mini & 78.2\% & 94.5\% & 86.1\% & 3.63\\
Zero-Shot Qwen
& Gemini 3.6 Flash & 79.1\% & 93.8\% & 85.4\% & 3.57\\
\midrule
Probe + Caption
& GPT-5.4 mini & 43.2\% & 71.3\% & 94.4\% & 2.85\\
Probe + Caption
& Gemini 3.6 Flash & 44.0\% & 70.8\% & 93.9\% & 2.78\\
\midrule
\textbf{RBA}
& \textbf{GPT-5.4 mini}
& \textbf{91.4\%} & \textbf{96.3\%}
& \textbf{36.7\%} & \textbf{4.12}\\
\textbf{RBA}
& \textbf{Gemini 3.6 Flash}
& \textbf{92.0\%} & \textbf{95.9\%}
& \textbf{35.8\%} & \textbf{4.09}\\
\bottomrule
\end{tabular}
\end{table}
These results provide evidence of improved decision--explanation coupling. They do not establish that the generated explanation is a complete causal account of the model's internal computation.
% -------------------------------------------------------------
% Appendix E
% -------------------------------------------------------------
\section{Statistical Significance and Multi-Seed Robustness}
\label{app:statistics}
We conduct five independent runs with random seeds
\[
S\in\{42,100,2024,2026,3407\}.
\]
Across runs, we vary the LoRA initialization, mini-batch ordering, and dropout masks while keeping the dataset splits, pseudo-labeling procedure, training hyperparameters, prompt template, and evaluation protocol fixed.
\begin{table}[h]
\centering
\small
\setlength{\tabcolsep}{8pt}
\renewcommand{\arraystretch}{1.1}
\caption{Multi-seed evaluation over five independent runs. Metrics are reported in percent. The base readout is deterministic under the fixed prompt and greedy decoding protocol.}
\label{tab:app_multiseed}
\begin{tabular}{llccc}
\toprule
\textbf{Dataset} & \textbf{Metric} & \textbf{Base Readout} & \textbf{RBA} & \textbf{Absolute Gain} \\
\midrule
\multirow{2}{*}{XD-Violence} 
& Frame-AP  & 72.86 & \textbf{83.02 $\pm$ 0.17} & +10.16 $\pm$ 0.17 \\
& Frame-AUC & 90.37 & \textbf{94.30 $\pm$ 0.11} & +3.93 $\pm$ 0.11 \\
\midrule
UCF-Crime   & Frame-AUC & 77.14 & \textbf{87.01 $\pm$ 0.19} & +9.87 $\pm$ 0.19 \\
\midrule
UBnormal    & Frame-AUC & 72.90 & \textbf{76.84 $\pm$ 0.22} & +3.94 $\pm$ 0.22 \\
\bottomrule
\end{tabular}
\end{table}
The standard deviations remain within $0.22$ percentage points for all reported metrics. Two-sided one-sample $t$-tests on the per-seed improvements over the deterministic base readout yield $p<10^{-5}$ for all evaluated dataset--metric pairs.
These results indicate that the observed improvements are substantially larger than the variation caused by LoRA initialization, mini-batch ordering, and stochastic dropout. The multi-seed results also support the stability of the transition-layer placement and the default rank-$16$, margin-$2.0$ configuration.
% -------------------------------------------------------------
% Appendix F
% -------------------------------------------------------------
\section{Quantitative Case Studies of Prediction Correction}
\label{app:quantitative_cases}
This section provides two representative case groups to quantify how RBA
corrects the native readout of the frozen Qwen3.5 backbone. Each group
contains one normal video and one abnormal video. We compare the anomaly
scores produced by the zero-shot native readout and RBA at representative
frames, using $0.5$ as the decision threshold.
For a clip with ground-truth label $y\in\{0,1\}$ and anomaly score $s$, we
define the signed decision margin as
\begin{equation}
m(s,y)
=
(2y-1)(s-0.5).
\label{eq:app_signed_margin}
\end{equation}
A positive margin denotes a correct prediction, while a negative margin
denotes an incorrect prediction. The margin improvement introduced by RBA is
\begin{equation}
\Delta m
=
m(s_{\mathrm{RBA}},y)
-
m(s_{\mathrm{ZS}},y),
\label{eq:app_margin_improvement}
\end{equation}
where $s_{\mathrm{ZS}}$ and $s_{\mathrm{RBA}}$ are the zero-shot and
RBA anomaly scores, respectively.
\subsection{Case Group I: Traffic and Outdoor Crowd Scenes}
\label{app:case_group_one}
\begin{figure}[!t]
    \centering
    \includegraphics[width=0.98\linewidth]{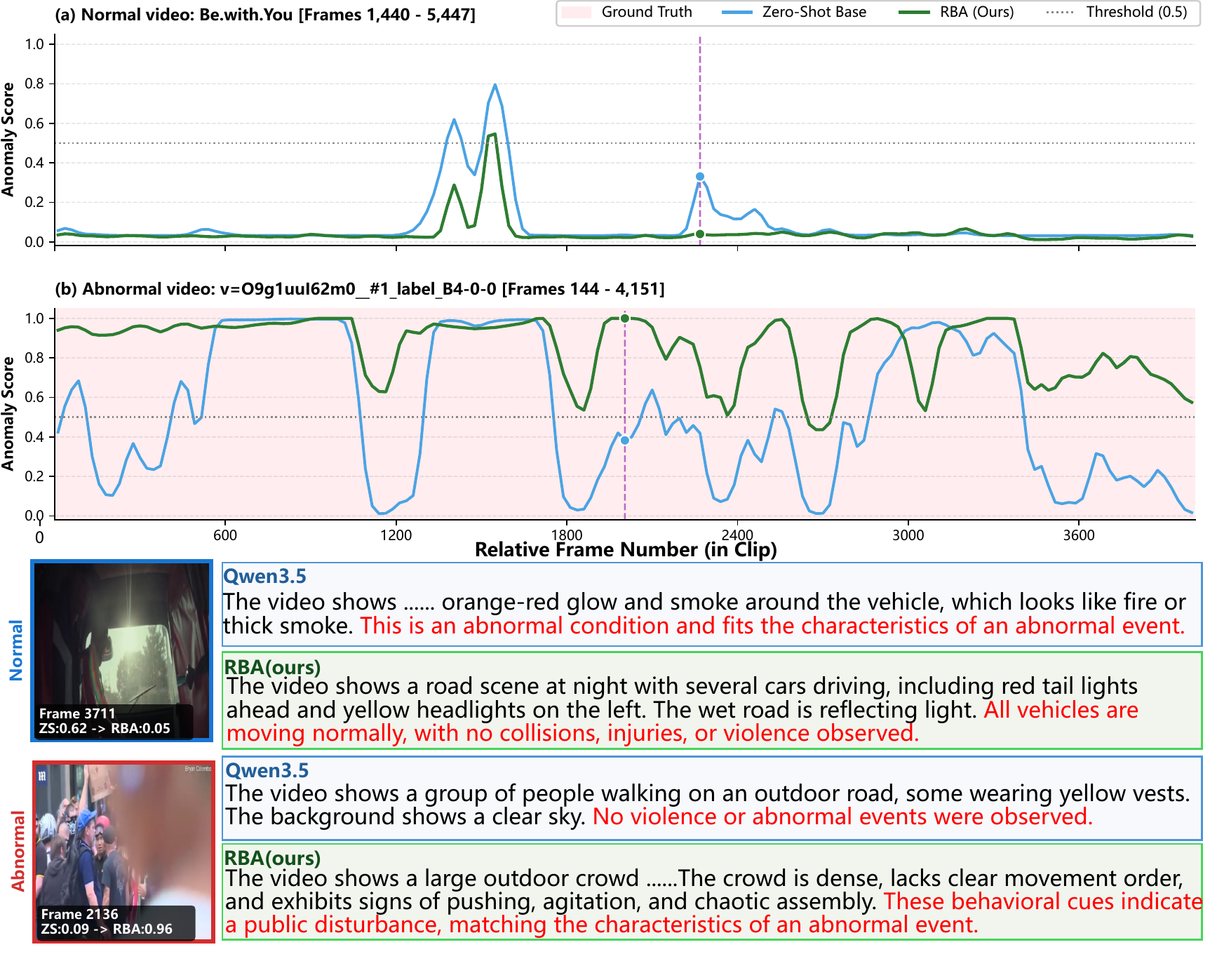}
    \caption{
    Quantitative and qualitative comparison for Case Group I.
    Left: temporal anomaly-score trajectories produced by the zero-shot
    Qwen3.5 readout and RBA, together with frame-level ground truth and the
    decision threshold.
    Right: explanations generated at representative frames.
    RBA suppresses a false alarm in the normal traffic video
    ($0.62\rightarrow0.05$) and recovers a missed anomaly in the abnormal
    crowd video ($0.09\rightarrow0.96$).
    }
    \label{fig:app_case_group_one}
\end{figure}
Figure~\ref{fig:app_case_group_one}(a) presents a normal nighttime traffic
video. At Frame~3711, the zero-shot readout assigns an anomaly score of
$0.62$, incorrectly exceeding the threshold. The prediction is associated
with visually salient but non-anomalous cues, including vehicle lights,
reflections on a wet road, and an orange-red glow. RBA reduces the score to
$0.05$, producing the correct normal prediction.
For this normal example, the signed margin changes from
\begin{equation}
m(0.62,0)=-0.12
\quad\text{to}\quad
m(0.05,0)=0.45,
\end{equation}
corresponding to a margin improvement of $\Delta m=0.57$.
Figure~\ref{fig:app_case_group_one}(b) shows an abnormal outdoor crowd
scene. The zero-shot model assigns only $0.09$ at Frame~2136 and therefore
misses the anomaly. RBA increases the score to $0.96$, correctly identifying
the event as abnormal. The corresponding signed margin changes from
\begin{equation}
m(0.09,1)=-0.41
\quad\text{to}\quad
m(0.96,1)=0.46,
\end{equation}
yielding $\Delta m=0.87$.
The generated explanations change consistently with the corrected
decisions. In the normal traffic example, RBA recognizes that the observed
lights and road reflections do not indicate collision, injury, or violence.
In the abnormal crowd example, RBA focuses on the dense and disorderly crowd
behavior rather than interpreting the scene as an ordinary gathering.
\FloatBarrier
\subsection{Case Group II: Sports and Dense Crowd Scenes}
\label{app:case_group_two}
\begin{figure}[!t]
    \centering
    \includegraphics[width=0.98\linewidth]{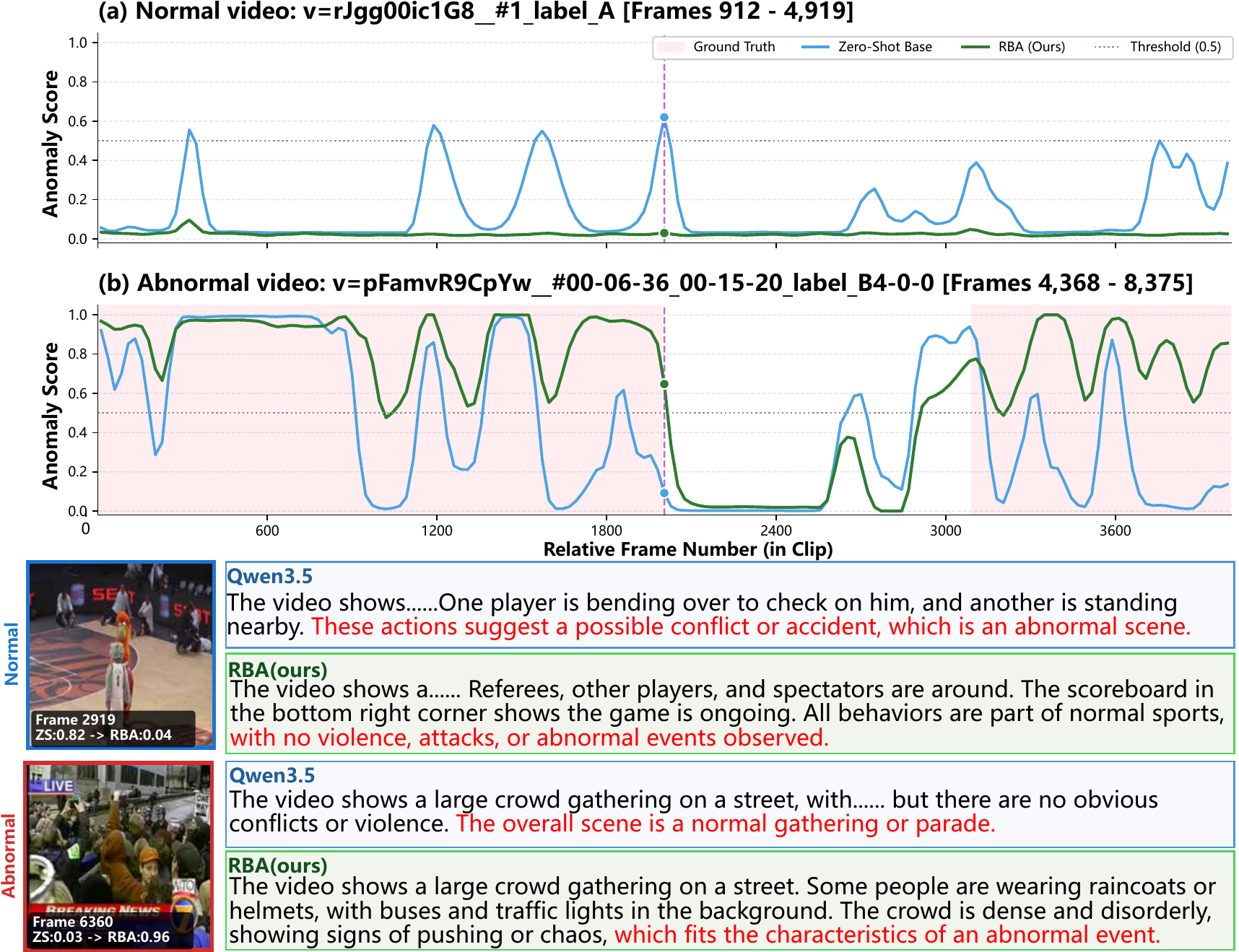}
    \caption{
    Quantitative and qualitative comparison for Case Group II.
    RBA corrects a false positive in a normal sports video
    ($0.82\rightarrow0.04$) and a false negative in an abnormal dense-crowd
    video ($0.03\rightarrow0.96$). The generated explanations are shown
    alongside the temporal score trajectories.
    }
    \label{fig:app_case_group_two}
\end{figure}
Figure~\ref{fig:app_case_group_two}(a) presents a normal sports video. At
Frame~2919, the zero-shot model produces an anomaly score of $0.82$,
apparently treating a player bending over and nearby people as evidence of a
possible conflict or accident. RBA lowers the score to $0.04$ and correctly
recognizes that the players, referees, spectators, and ongoing scoreboard
belong to a normal sports context.
The signed margin for this example changes from
\begin{equation}
m(0.82,0)=-0.32
\quad\text{to}\quad
m(0.04,0)=0.46,
\end{equation}
which gives a margin improvement of $\Delta m=0.78$.
Figure~\ref{fig:app_case_group_two}(b) presents an abnormal dense-crowd
scene. The zero-shot model assigns a score of $0.03$ at Frame~6360 and
describes the event as a normal gathering or parade. In contrast, RBA
increases the anomaly score to $0.96$. Its explanation identifies the dense
and disorderly crowd, protective clothing, and signs of pushing or chaotic
movement as evidence of an abnormal event.
The signed margin changes from
\begin{equation}
m(0.03,1)=-0.47
\quad\text{to}\quad
m(0.96,1)=0.46,
\end{equation}
resulting in the largest observed improvement among these examples,
$\Delta m=0.93$.
\subsection{Quantitative Summary}
\label{app:case_quantitative_summary}
Table~\ref{tab:app_case_summary} summarizes the representative-frame
results. All four zero-shot decisions are initially located on the incorrect
side of the $0.5$ threshold. RBA moves every example to the correct side,
with an average signed-margin improvement of
\begin{equation}
\overline{\Delta m}
=
\frac{0.57+0.87+0.78+0.93}{4}
=
0.7875.
\end{equation}
\begin{table}[!t]
    \centering
    \small
    \setlength{\tabcolsep}{4.5pt}
    \caption{
    Quantitative summary of the representative cases. ZS denotes the
    zero-shot native readout. A positive signed margin indicates a correct
    prediction under the $0.5$ threshold.
    }
    \label{tab:app_case_summary}
    \resizebox{\linewidth}{!}{
    \begin{tabular}{cllcccccc}
        \toprule
        \textbf{Group}
        & \textbf{Scene}
        & \textbf{GT}
        & \textbf{Frame}
        & \textbf{ZS Score}
        & \textbf{RBA Score}
        & \textbf{ZS Margin}
        & \textbf{RBA Margin}
        & $\boldsymbol{\Delta m}$ \\
        \midrule
        I
        & Nighttime traffic
        & Normal
        & 3711
        & 0.62
        & \textbf{0.05}
        & $-0.12$
        & $\mathbf{+0.45}$
        & $\mathbf{+0.57}$ \\
        I
        & Outdoor crowd
        & Abnormal
        & 2136
        & 0.09
        & \textbf{0.96}
        & $-0.41$
        & $\mathbf{+0.46}$
        & $\mathbf{+0.87}$ \\
        II
        & Sports activity
        & Normal
        & 2919
        & 0.82
        & \textbf{0.04}
        & $-0.32$
        & $\mathbf{+0.46}$
        & $\mathbf{+0.78}$ \\
        II
        & Dense crowd
        & Abnormal
        & 6360
        & 0.03
        & \textbf{0.96}
        & $-0.47$
        & $\mathbf{+0.46}$
        & $\mathbf{+0.93}$ \\
        \midrule
        \multicolumn{6}{r}{\textbf{Mean}}
        & $-0.33$
        & $\mathbf{+0.46}$
        & $\mathbf{+0.79}$ \\
        \bottomrule
    \end{tabular}
    }
\end{table}
Across the four representative frames, the zero-shot readout obtains
$0/4$ correct decisions, whereas RBA obtains $4/4$. For the two normal
examples, the mean anomaly score decreases from
\begin{equation}
\frac{0.62+0.82}{2}=0.72
\quad\text{to}\quad
\frac{0.05+0.04}{2}=0.045.
\end{equation}
For the two abnormal examples, the mean anomaly score increases from
\begin{equation}
\frac{0.09+0.03}{2}=0.06
\quad\text{to}\quad
\frac{0.96+0.96}{2}=0.96.
\end{equation}
These cases illustrate two complementary effects of representation--behavior
alignment. First, RBA suppresses false positives caused by visually salient
but contextually normal patterns, such as road reflections and ordinary
sports interactions. Second, it recovers false negatives in crowded scenes
whose abnormality depends on contextual and behavioral evidence rather than
isolated objects. The simultaneous correction of anomaly scores and generated
explanations is consistent with the intended alignment of task-discriminative
representations with the model's native generative readout.

\end{document}